\pdfoutput=1
\documentclass[11pt]{article}

\usepackage[hyperref]{style/emnlp2021}

\usepackage{times}
\usepackage{soul}
\usepackage{latexsym}

\usepackage[T1]{fontenc}

\usepackage[utf8]{inputenc}

\usepackage{microtype}

\newif\ifcomment\commentfalse

\newif\ifdraft\draftfalse

\usepackage[a-1b]{pdfx}

\usepackage{framed}
\usepackage{mdwlist}
\usepackage{siunitx}
\usepackage{latexsym}
\usepackage{colortbl}
\usepackage{xcolor}
\usepackage{nicefrac}
\usepackage{booktabs}
\usepackage{fnpct}
\usepackage{amsfonts}
\usepackage[T1]{fontenc}
\usepackage{bold-extra}
\usepackage{amsmath}
\usepackage{amssymb}
\usepackage{bm}
\usepackage{graphicx}
\usepackage{mathtools}
\usepackage{microtype}
\usepackage{multirow}
\usepackage{multicol}
\usepackage{todonotes}
\usepackage{xpatch}
\usepackage{latexsym,comment}
\usepackage[normalem]{ulem}

\newcommand*{\missingreference}{{\Huge \colorbox{red}{?reference?}}}
\newcommand*{\missingcitation}{{\Huge \colorbox{red}{?citation?}}}

\makeatletter
\xpatchcmd{\@setref}{\bfseries}{\missingreference}{}{}
\def\@citex[#1]#2{\leavevmode
    \let\@citea\@empty
    \@cite{\@for\@citeb:=#2\do
        {\@citea\def\@citea{,\penalty\@m\ }%
            \edef\@citeb{\expandafter\@firstofone\@citeb\@empty}%
            \if@filesw\immediate\write\@auxout{\string\citation{\@citeb}}\fi
            \@ifundefined{b@\@citeb}{\hbox{\reset@font\missingcitation}%
                \G@refundefinedtrue
                \@latex@warning
                {Citation `\@citeb' on page \thepage \space undefined}}%
            {\@cite@ofmt{\csname b@\@citeb\endcsname}}}}{#1}}
\makeatother

\newcommand{\gem}[1]{\mbox{\textsc{gem}}}
\newcommand{\abr}[1]{\textsc{#1}}
\newcommand{\camelabr}[2]{{\small #1}{\textsc{#2}}}

\newcommand{\emaillink}[1]{{\small \href{mailto://#1}{\texttt{#1}}}}

\newcommand{\hidetext}[1]{}
\newcommand{\ignore}[1]{}

\ifcomment
    \newcommand{\pinaforecomment}[3]{\colorbox{#1}{\parbox{.8\linewidth}{#2: #3}}}
\newcommand{\prtodo}[1]{\todo[backgroundcolor=lightblue,linecolor=lightblue,author=Pedro]{#1}}
\newcommand{\prtodoi}[1]{\todo[inline,backgroundcolor=lightblue,linecolor=lightblue,author=Pedro]{#1}}
\else
    \newcommand{\pinaforecomment}[3]{}
\newcommand{\prtodo}[1]{}
\newcommand{\prtodoi}[1]{}
\fi

\newcommand{\smallurl}[1]{ \begin{tiny}\url{#1}\end{tiny}}

\definecolor{lightblue}{HTML}{3cc7ea}
\definecolor{CUgold}{HTML}{CFB87C}
\definecolor{grey}{rgb}{0.95,0.95,0.95}
\definecolor{ceil}{rgb}{0.57, 0.63, 0.81}
\definecolor{UMDred}{HTML}{ed1c24}
\definecolor{UMDyellow}{HTML}{ffc20e}
\definecolor{GoogleBlue}{HTML}{4285f4}
\definecolor{GoogleRed}{HTML}{ea4335}
\definecolor{GoogleYellow}{HTML}{fbbc04}
\definecolor{GoogleGreen}{HTML}{34a853}

\newcommand{\qb}[0]{\abr{qb}}
\newcommand{\qa}[0]{\abr{qa}}

\newcommand{\triviaqa}{\camelabr{Trivia}{qa}}

\newcommand{\squad}{\textsc{sq}{\small u}\textsc{ad}}
\newcommand{\nq}[0]{\abr{nq}}

\newcommand{\bert}{\abr{bert}}

\usepackage{tikz}
\usepackage{collcell}

\newcommand*{\MinNumber}{0}%
\newcommand*{\MidNumber}{100} %
\newcommand*{\MaxNumber}{200}%

\newcommand{\ApplyGradient}[1]{%
        \ifdim #1 pt > \MidNumber pt
            \pgfmathsetmacro{\PercentColor}{max(min(100.0*(#1 - \MidNumber)/(\MaxNumber-\MidNumber),100.0),0.00)} %
            \colorbox{red!\PercentColor!UMDYellow}{#1}
        \else
            \pgfmathsetmacro{\PercentColor}{max(min(100.0*(\MidNumber - #1)/(\MidNumber-\MinNumber),100.0),0.00)} %
            \colorbox{white!\PercentColor!GoogleYellow}{#1}
        \fi
}

\newcommand{\cbcol}[1]{\multicolumn{1}{c}{\textbf{#1}}}
\newcommand{\cicol}[1]{\multicolumn{1}{c}{\textit{#1}}}

\newcolumntype{R}{>{\collectcell\ApplyGradient}c<{\endcollectcell}}

\usepackage{microtype}

\title{Toward Deconfounding the Influence of Entity Demographics for Question Answering Accuracy}

\newcommand{\entity}[1]{\ul{#1}}
\newcommand{\answer}[1]{\textsl{#1}}

\newcommand{\demovalue}[1]{\textit{#1}}
\newcommand{\lrfeature}[1]{\begin{small}\textbf{\texttt{#1}}\end{small}}
\newcommand{\feature}[1]{\begin{small}{\texttt{#1}}\end{small}}

\newcommand{\wikidata}[0]{Wikidata}

\newcommand{\democol}{characteristic}
\newcommand{\demorow}{value}
\newcommand{\demosubset}{subset}

\newcommand{\film}{\demovalue{Film/TV}}
\newcommand{\music}{\demovalue{Music}}
\newcommand{\sports}{\demovalue{Sports}}

\newcommand{\lasso}[0]{{\abr{lasso}}}

\definecolor{lightgrey}{rgb}{.8,.8,.8}

\author{Maharshi Gor\thanks{\, \, Work completed while at Google Research} \\ \abr{CS} \\ University of Maryland \\ \emaillink{mgor@cs.umd.edu}
 \And Kellie Webster \\ Google Research, New York \\ \emaillink{websterk@google.com}
 \And Jordan Boyd-Graber$^{*}$ \\ CS, UMIACS, iSchool, LCS \\ University of Maryland \\ \emaillink{jbg@umiacs.umd.edu}}

\begin{document}

\maketitle
\begin{abstract}
    
The goal of question answering (\abr{qa}) is to answer \emph{any} question.
However, major \abr{qa} datasets have skewed distributions over gender, profession, and nationality.
Despite that skew, model accuracy analysis reveals little evidence that accuracy is lower for people based on gender or nationality; instead, there is more variation on professions (question topic).
But \abr{qa}'s lack of representation could itself hide evidence of bias, necessitating \abr{qa} datasets that better represent global diversity.
%
%

\end{abstract}

\section{Introduction}

Question answering (\abr{qa}) systems have impressive recent victories---beating trivia masters~\cite{ferruci-10} and superhuman reading~\cite{najberg-18}---but these triumphs hold only if they \emph{generalize}; \abr{qa} systems should be able to answer questions even if they do not look like training examples.
While other work (Section~\ref{sec:related}) focuses on demographic representation in \abr{nlp} resources, our focus is how well \abr{qa} models generalize across demographic \demosubset{}s.

After mapping mentions to a knowledge base (Section~\ref{sec:mapping}), we show existing \abr{qa} datasets lack diversity in the gender and national origin of the people mentioned: English-language \abr{qa} datasets mostly ask about \abr{us} men from a few professions  (Section~\ref{sec:distribution}).
This is problematic because most English speakers (and users of English \abr{qa} systems) are not from the \abr{us} or \abr{uk}.
Moreover, multilingual \abr{qa} datasets are often \emph{translated} from English datasets~\cite{lewis-etal-2020-mlqa,
  artetxe2019xquad}.
However, no work has verified that \abr{qa} systems generalize to infrequent demographic groups.

Section~\ref{sec:accuracy} investigates whether statistical tests reveal patterns on demographic subgroups.  
Despite skewed distributions, accuracy is not correlated with gender or nationality, though it is with professional field.
For instance, Natural Questions~\cite[\abr{nq}]{kwiatkowski-19} systems do well with entertainers but poorly with scientists, which are handled well in \triviaqa{}.
However, absence of evidence is not evidence of absence (Section~\ref{sec:conclusion}), and existing \abr{qa} datasets are not yet diverse enough to vet \abr{qa}'s generalization.

\begin{table}[t]
    \centering
    \setlength{\tabcolsep}{4pt} 
    \resizebox{\linewidth}{!}{%
        \begin{tabular}{l *{8}{R}}
        \multirow{2}{*}{\textbf{Entity Type}} 
            & \multicolumn{2}{c}{\textbf{\nq}} 
            & \multicolumn{2}{c}{\textbf{\qb}}  
            & \multicolumn{2}{c}{\textbf{\squad}}  
            & \multicolumn{2}{c}{\textbf{\triviaqa}} \\ 
        
        & \cbcol{Train} & \cbcol{Dev} & \cbcol{Train} & \cbcol{Dev} & \cbcol{Train} & \cbcol{Dev} & \cbcol{Train} & \cbcol{Dev}\\ 
        
        \toprule
        
        \textbf{Person} 
            & 32.14 &  32.27
            & 58.49 &  57.18
            & 14.56 &   9.56 
            & 40.89 &  40.69 \\
        
        No entity       
            & 21.97 &  22.61 
            & 15.88 &  20.13 
            & 37.27 &  47.95 
            & 14.85 &  14.93 \\        
        
        Location        
            & 28.42 &  27.82 
            & 34.50 &  35.11 
            & 33.52 &  29.34 
            & 44.32 &  44.10 \\
        
        Work of art     
            & 17.31 &  16.19 
            & 27.98 &  28.16 
            & 3.15  &   1.15 
            & 17.53 &  17.76 \\
        
        Other           
            &   2.66 & 2.70 
            &   9.83 & 9.88 
            &   5.08 & 3.10 
            &  14.38 & 14.61 \\

        Organization    
            &  17.73 & 18.05 
            &  20.37 & 17.78 
            &  21.34 & 19.14 
            &  22.15 & 21.14 \\
        
        Event           
            &   8.15 &   8.89 
            &   8.87 &   8.75 
            &   3.16 &   3.01 
            &   8.59 &   8.73 \\
        
        Product         
            &   0.85 &   1.06 
            &   0.88 &   0.63 
            &   1.39 &   0.33 
            &   3.99 &   3.99 \\
        
        \rowcolor{lightgrey}
        Total Examples  
            & \cicol{106926} & \cicol{2631}
            & \cicol{112927} & \cicol{2216}
            & \cicol{130319} & \cicol{11873}
            & \cicol{87622}  &  \cicol{11313} \\
        \bottomrule
        \end{tabular}
    }
\caption{Coverage (\% of examples) of entity-types in \abr{qa} datasets. Since examples can mention more than one entity, columns can sum to $>$100\%. Most datasets except \squad{} are rich in people.}
\label{tab:entity-distribution}
\end{table}





\section{Mapping Questions to Entities}
\label{sec:mapping}

We analyze four \abr{qa} tasks: \nq{},\footnote{For \nq{}, we only consider questions with short answers.} \squad~\cite{rajpurkar-16}, \qb~\cite{boyd-graber-12} and \triviaqa~\cite{joshi-17}.
Google \abr{cloud-nl}\footnote{\smallurl{https://cloud.google.com/natural-language/docs/analyzing-entities}} finds and links entity mentions in \qa{} examples.\footnote{We analyze the dev fold, which is {\bf consistent with the training fold} (Table~\ref{tab:entity-distribution} and~\ref{tab:demographics}), as we examine accuracy.}

\subsection{Focus on \emph{People}}
\label{subsec:people}
Many entities appear in examples (Table~\ref{tab:entity-distribution}) but \emph{people} form a majority in our \abr{qa} tasks (except \squad{}). Existing work in \abr{ai} fairness focuses on disparate impacts on people, and model behaviors are prone to harm especially when it comes to \emph{people}; hence, our primary intent is to understand how demographic \democol{}s of ``people'' correlate with model correctness.

The people asked about in a question can be in the answer---``who founded Sikhism?'' (A: \answer{\entity{Guru Nanak}}), in the question---``what did \entity{Clara Barton} found?'' (A: American Red Cross), or the title of the source document---``what play featuring General Uzi premiered in Lagos in 2001?'' (A: \answer{King Baabu}) is in the page on \entity{Wole Soyinka}.
We search until we find an entity: first in the answer, then the question if no entity is found in the answer, and finally the document title.

Demographics are a natural way to categorize these entities and we consider the high-coverage demographic {\bf \democol{}s} from \wikidata{}.\footnote{\smallurl{https://www.wikidata.org/wiki/Wikidata:Database_download}}
Given an entity, Wikidata has good coverage for all datasets: gender ($>99\%$ ), nationality ($>93\%$), and profession ($>94\%$).
For each \democol{}, we use the knowledge base to extract the specific {\bf \demorow{}} for a person (e.g., the \demorow{} ``poet'' for the \democol{} ``profession'').
However, the \demorow{}s defined by \wikidata{} have inconsistent granularity, so we collapse near-equivalent \demorow{}s (E.g., ``writer'', ``author'', ``poet'', etc. See Appendix~\ref{appendix:country-collapse}--\ref{appendix:professions-collapse} for an exhaustive list).
For questions with multiple values (where multiple entities appear in the answer, or a single entity has multiple \demorow{}s), we create a new value concatenating them together. 
An `others’ \demorow{} subsumes \demorow{}s with fewer than fifteen examples; people without a \demorow{} become `not found' for that \democol{}.

%

\label{entity-linking-procedure}

%
%
%
%
%

\label{entity-linking-validation}

Three authors manually verify entity assignments by vetting fifty random questions from each dataset. Questions with at least one entity had near-perfect 96\% inter-annotator agreement for \abr{cloud-nl}'s annotations, while for questions where \abr{cloud-nl} didn't find any entity, agreement is 98\%. 
%
%
%
Some errors were benign: incorrect entities sometimes retain correct demographic \demorow{}s; e.g., \entity{Elizabeth~II} instead of \entity{Elizabeth~I}. Other times, coarse-grained nationality ignores nuance, such as the distinction between \emph{Greece} and \emph{Ancient Greece}.

\subsection{Who is in Questions?}
\label{sec:distribution}

Our demographic analysis reveals skews in all datasets, reflecting differences in task focus (Table~\ref{tab:demographics}).
\nq{} is sourced from search queries and skews toward popular culture.
\qb{} nominally reflects an undergraduate curriculum and captures more ``academic'' knowledge.
%
\triviaqa{} is popular trivia, and \squad{} reflects Wikipedia articles.

Across all datasets, men are asked about more than women, and the \abr{us} is the subject of the majority of questions except in \triviaqa{}, where the plurality of questions are about the \abr{uk}.
\nq{} has the highest coverage of women through its focus on entertainment (\film{}, \music{} and \sports{}).

\begin{table}[t]
    \centering
    \setlength{\tabcolsep}{3pt} 
    \resizebox{\linewidth}{!}{%
        \begin{tabular}{cl*{8}{R}}
        \hspace{15pt} & \multirow{2}{*}{\textbf{Value}} 
            & \multicolumn{2}{c}{\textbf{\nq}} 
            & \multicolumn{2}{c}{\textbf{\qb}}  
            & \multicolumn{2}{c}{\textbf{\squad}}  
            & \multicolumn{2}{c}{\textbf{\triviaqa}} \\ 
        & 
            & \cbcol{Train} & \cbcol{Dev}
            & \cbcol{Train} & \cbcol{Dev} 
            & \cbcol{Train} & \cbcol{Dev} 
            & \cbcol{Train} & \cbcol{Dev}\\         
        \toprule
        \multirow{3}{*}{\rotatebox[origin=c]{90}{\textbf{Gender}}}
            & Male    
                &  75.67 &  76.33 
                &  91.77 &  91.63 
                &  87.82 &  95.15 
                &  83.76 &  83.32 \\
               
            & Female          
                &  27.47 &  27.56 
                &  10.29 &   9.87 
                &  13.44 &   5.20 
                &  20.54 &  20.29 \\
                
            & No Gender
                &   0.31  &   0.47
        		&   0.35  &   0.39
        		&   0.27  &   0.00
        		&   0.30  &   0.35 \\
        \midrule
        \multirow{6}{*}{\rotatebox[origin=c]{90}{\textbf{Country}}}
            & US
                & 59.62	& 58.66
        		& 29.70	& 26.28
        		& 32.74	& 24.93
        		& 31.32	& 30.91
                \\
            
            & UK
        		& 15.76	& 15.78
        		& 17.92	& 17.68
        		& 19.66	& 16.83
        		& 41.92	& 41.32
        		\\
            
            & France
        		& 1.79	& 1.18
        		& 10.06	& 10.34
        		& 7.76	& 10.57
        		& 4.37	& 4.84
        		\\
            
            & Italy
				& 1.83	& 1.88
				& 8.07	& 10.50
				& 9.00	& 3.88
				& 3.75	& 3.48
				\\
            & Germany        
				& 1.52	& 2.12
				& 7.21	& 6.71
				& 4.77	& 6.61
				& 3.01	& 3.00
				\\

            & No country 
				& 4.82	& 4.36
				& 7.12	& 6.79
				& 3.48	& 2.56
				& 6.19	& 6.10
				\\

        \midrule
        \multirow{7}{*}{\rotatebox[origin=c]{90}{\textbf{Professional Field}}}
            & Film/TV
				& 39.19	& 37.93
				& 3.16	& 1.89
				& 10.72	& 1.32
				& 20.64	& 20.75
				\\
            
            & Writing 
				& 7.40	& 6.95
				& 36.62	& 36.39
				& 10.46	& 6.70
				& 18.41	& 18.05
				\\
				
            & Politics 
				& 11.98	& 10.84
				& 24.02	& 24.86
				& 36.97	& 46.61
				& 21.18	& 20.72
				\\
				
            & Science/Tech
				& 3.61	& 4.71
				& 8.93	& 7.50
				& 13.67	& 29.60
				& 5.43	& 5.54
				\\
				
            & Music  
				& 24.08	& 23.44
				& 9.51	& 10.73
				& 13.56	& 2.64
				& 16.46	& 17.38
				\\
				
            & Sports
				& 13.69	& 16.49
				& 1.31	& 0.32
				& 2.99	& 1.76
				& 11.03	& 11.19
				\\
				
            & Stage
				& 15.67	& 14.84
				& 1.49	& 1.03
				& 1.21	& 1.23
				& 10.89	& 10.28
				\\
			
			& Artist
				& 1.84	& 2.47
				& 9.17	& 9.00
				& 4.58	& 2.38
				& 5.75	& 5.93
				\\
				
        \bottomrule
        \end{tabular}
    }
    \caption{\label{table-dataset-demographics} Coverage (\% of examples) of demographic \demorow{}s across examples with people in \abr{qa} datasets. Men dominate, as do Americans.}
    \label{tab:demographics}
\end{table}

\begin{figure*}[t]
  \includegraphics[width=\linewidth]{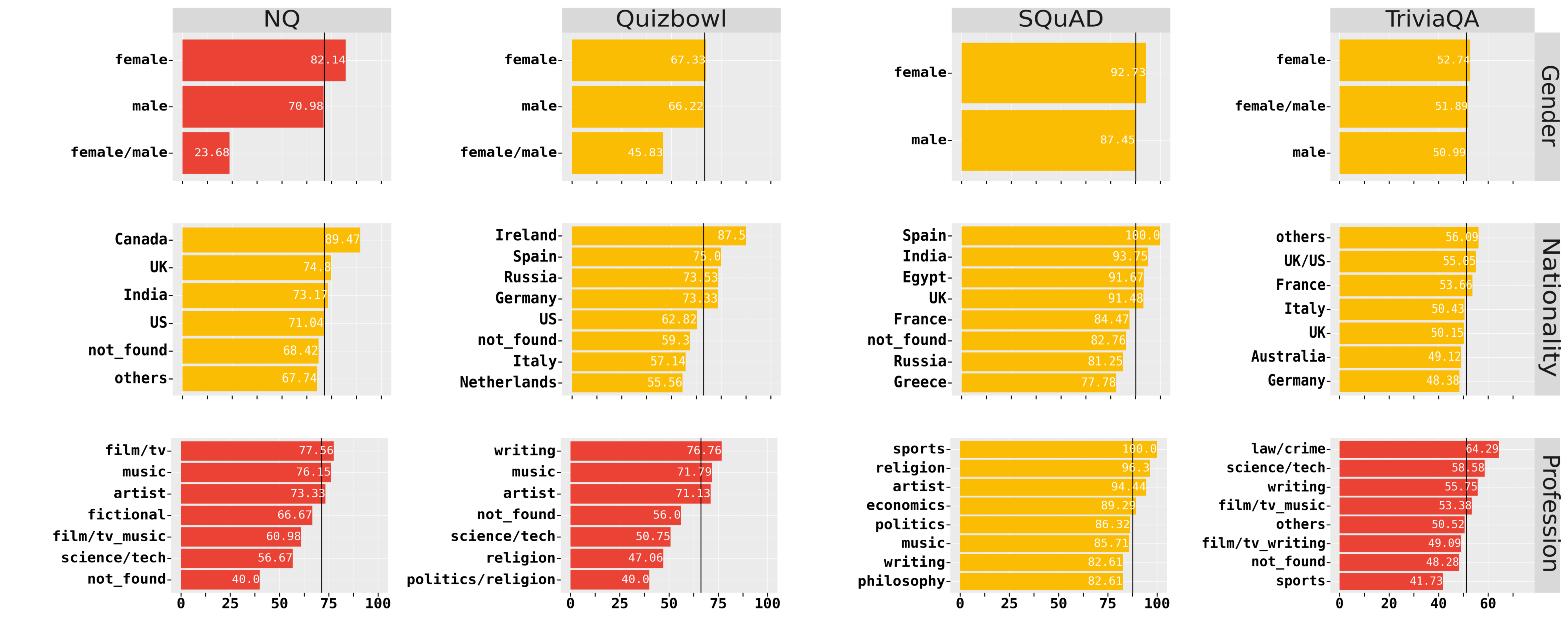}
  \caption{
  Accuracies split by demographic \demosubset{}s in our \abr{qa} datasets' dev fold for all three \democol{}s compared to average accuracy (vertical line). For each dataset, we only consider examples that has a mention of a person-entity in either the answer, question or document title.
  Individual plots correspond to a $\chi^2$ test on whether demographic \demorow{}s and accuracy are independent (Section~\ref{subsec:chi-squared-test}) with the significant \democol{}s highlighted in red ($p$-value < 0.0167).}
  \label{fig:accuracies}
\end{figure*} 
\section{What Questions can \abr{qa} Answer?}
\label{sec:accuracy}

\qa{} datasets have different representations of demographic \democol{}s; 
is this focus benign or do these differences carry through to model accuracy?

We analyze a \abr{sota} system for each of our four tasks.
For \nq{} and \squad{}, we use a fine-tuned \bert{}~\cite{alberti2019bert} with curated training data (e.g., downsample questions without answers and split documents into multiple training instances). For the open-domain \triviaqa{} task, we use \abr{orqa}~\cite{lee2019latent} that uses \bert{}-based reader and retriever components. Finally, for \qb{}, we use the competition winner from \citet{wallace-19}, a \abr{bert}-based reranker of a \abr{tf-idf} retriever.
Accuracy (exact-match) and average F1 are both common \abr{qa} metrics~\cite{rajpurkar-16}. Since both are related and some statistical tests require binary scores, we focus on exact-match.

Rather than focus on aggregate accuracy, we focus on demographic \demosubset{}s' accuracy (Figure~\ref{fig:accuracies}). 
For instance, while 66.2\% of questions about people are correct in \qb{}, the number is lower for the Dutch (\demovalue{Netherlands}) (55.6\%) and higher for  \demovalue{Ireland} (87.5\%).
Unsurprisingly, accuracy is consistently low on the `not\_found' \demosubset{}, where \wikidata{} lacks a person's demographic \demorow{}.
%

Are the differences we observe across strata significant?
We probe this in two ways: using $\chi^2$ testing~\cite{plackett1983karl} to see \emph{if} trends exist and using logistic regression to explore those that do.

\subsection{Do Demographic Values Affect Accuracy?}
\label{subsec:chi-squared-test}
The $\chi^2$ test is a non-parametric test of whether two variables are independent. 
To see if accuracy and \democol{}s are independent, we apply a $\chi^2$ test to a $n \times 2$ contingency table with $n$ rows representing the frequency of that \democol{}'s \demosubset{}s contingent on whether the model prediction is correct or not (Table~\ref{tab:contingency}).
If we reject the null with a Bonferroni correction~\cite[divide the $p$-value threshold by three, as we have multiple tests for each dataset]{holm-79}, that suggests possible relationships:
gender in \nq{} ($p=$\SI{2.36e-12}), and professional field in \nq{} ($p=0.0142$), \qb{} ($p=$\SI{2.34e-07}) and \triviaqa{} ($p=0.0092$). 
However, we find no significant relationship between nationality and accuracy in any dataset.
%

While $\chi^{2}$ identifies \emph{which} \democol{}s impact model accuracy, it does not characterize \emph{how}.
For instance, $\chi^2$  indicates \nq{}'s gender is significant, but is this because accuracy is higher for women, or because the presence of both genders in examples lowers the accuracy?

\begin{table}[t]
    \centering
    \setlength{\tabcolsep}{6pt} 
    \resizebox{\linewidth}{!}{%
        \begin{tabular}{lccc}
        \textbf{Gender} & \textbf{\# Correct} & \textbf{\# Incorrect} & \textbf{Accuracy} \\
        
        \toprule
        
        \textbf{\lrfeature{male}}           & \cellcolor[HTML]{C0C0C0}433   & \cellcolor[HTML]{C0C0C0}177   & 70.98\% \\
        \textbf{\lrfeature{female}}         & \cellcolor[HTML]{C0C0C0}161   & \cellcolor[HTML]{C0C0C0}35    & 82.14\% \\
        \textbf{\lrfeature{female/male}}   & \cellcolor[HTML]{C0C0C0}9     & \cellcolor[HTML]{C0C0C0}29    & 23.68\% \\
        \bottomrule
        \end{tabular}
    }
\caption{Illustration of $n \times 2$ contingency table for the $\chi^2$-test: \textbf{Gender} in \textbf{\nq{}}, with $n = 3$ values: \lrfeature{male}, \lrfeature{female} and \lrfeature{female/male}. Female entities have a higher accuracy (82\%) than male (71\%). With two degrees of freedom and $\chi^2 = 53.55$, we get $p= $\SI{2.4e-12}, signaling significance.}
\label{tab:contingency}
\end{table}

\subsection{Exploration with Logistic Regression}
\label{subsec:logistic-regression}

\newcommand{\hfeat}[1]{\textbf{\lrfeature{#1}}}

Thus, we formulate a simple logistic regression: 
can an example's demographic \demorow{}s predict if a model answers correctly?
Logistic regression and related models are the workhorse for discovering and explaining the relationship between variables in history~\cite{mccloskey-87}, education~\cite{Linden-2013}, political science~\cite{poole-11}, and sports~\cite{glickman-99}.
Logistic regression is also a common tool in \abr{nlp}: to find linguistic constructs that allow determiner omission~\cite{kiss-10} or to understand how a scientific paper's attributes effect citations~\cite{yogatama-11}. 
Unlike model calibration~\cite{niculescu2005predicting}, whose goal it to maximize prediction accuracy, the goal here is \emph{explanation}.

%
We define binary features for demographic \demorow{}s of \democol{}s the $\chi^2$ test found significant (thus, \squad{}, the nationality \democol{}, and gender \democol{} for all but \nq{} are excluded).
For instance, a question about \entity{Abidali Neemuchwala} would have features for \hfeat{g\_male}, \hfeat{o\_executive} but zero for everything else.\footnote{Exhaustive list of demographic features in the Appendix.}
Real-valued features, \hfeat{multi\_entities} and \hfeat{multi\_answers}, capture the effect of multiple person-entities and multiple gold-answers (scaled with the base two logarithm). %
%

\definecolor{pos_color}{rgb}{1.0, 0.80392157, 0.1372549}
\definecolor{neg_color}{rgb}{0.81176471, 0.1254902 , 0.17647059}
\definecolor{ntr_color}{rgb}{0.0, 0.0, 0.0}

\newcommand{\postext}[1]{{\textcolor{GoogleBlue}{#1}}}
\newcommand{\negtext}[1]{{\textcolor{GoogleRed}{#1}}}
\newcommand{\blutext}[1]{{\textcolor{ntr_color}{#1}}}
\newcommand{\mystar}{{\fontfamily{lmr}\selectfont$\star$}}

\begin{table*}[t]
    \small
    \centering
    \setlength{\tabcolsep}{13pt} 
    \resizebox{\linewidth}{!}{%
        \begin{tabular}{llrcccc}
        \textbf{Dataset} & \textbf{Features}  & \textbf{Coef} & \textbf{SE} & \textbf{Wald (|W|)} & \multicolumn{2}{l}{\textbf{$\mathbb{P}_{Z\sim\mathcal{N}}(|Z| > |W|)$}} \\ 
        \toprule
\multirow{8}{*}{\parbox{2.0 cm}{\textbf{\nq{} \\  \\  \\ Model~Fit: \\ 78.09\%}}}
	 & \blutext{\lrfeature{\textbf{bias}}}
	 & \blutext{\textbf{+1.964}}
	 & \blutext{\textbf{0.727}}
	 & \blutext{\textbf{2.703}}
	 & \blutext{\textbf{0.0069}}
	 & \blutext{\textbf{\mystar\mystar\mystar}}
	
\\

	 & \negtext{\lrfeature{\textbf{multi\_answers}}}
	 & \negtext{\textbf{-1.893}}
	 & \negtext{\textbf{0.438}}
	 & \negtext{\textbf{4.327}}
	 & \negtext{\textbf{0.0000}}
	 & \negtext{\textbf{\mystar\mystar\mystar\mystar}}
	
\\

	 & \negtext{\lrfeature{\textbf{o\_not\_found}}}
	 & \negtext{\textbf{-1.112}}
	 & \negtext{\textbf{0.514}}
	 & \negtext{\textbf{2.163}}
	 & \negtext{\textbf{0.0305}}
	 & \negtext{\textbf{\mystar\mystar}}
	
\\

	 & \postext{\lrfeature{\textbf{t\_who}}}
	 & \postext{\textbf{+0.773}}
	 & \postext{\textbf{0.280}}
	 & \postext{\textbf{2.764}}
	 & \postext{\textbf{0.0057}}
	 & \postext{\textbf{\mystar\mystar\mystar}}
	
\\

	 & \negtext{\lrfeature{o\_science/tech}}
	 & \negtext{-0.715}
	 & \negtext{0.390}
	 & \negtext{1.832}
	 & \negtext{0.0670}
	 & \negtext{\textbf{\mystar}}
	
\\

	 & \negtext{\lrfeature{\textbf{multi\_entities}}}
	 & \negtext{\textbf{-0.678}}
	 & \negtext{\textbf{0.342}}
	 & \negtext{\textbf{1.979}}
	 & \negtext{\textbf{0.0478}}
	 & \negtext{\textbf{\mystar\mystar}}
	
\\

	 & \postext{\lrfeature{q\_sim}}
	 & \postext{+0.406}
	 & \postext{0.210}
	 & \postext{1.934}
	 & \postext{0.0531}
	 & \postext{\textbf{\mystar}}
	
\\

	 & \postext{\lrfeature{\textbf{e\_train\_count}}}
	 & \postext{\textbf{+0.353}}
	 & \postext{\textbf{0.178}}
	 & \postext{\textbf{1.979}}
	 & \postext{\textbf{0.0479}}
	 & \postext{\textbf{\mystar\mystar}}
	
\\

\midrule
\multirow{8}{*}{\parbox{2.0 cm}{\textbf{\qb{} \\  \\  \\ Model~Fit: \\ 71.90\%}}}
	 & \postext{\lrfeature{\textbf{e\_train\_count}}}
	 & \postext{\textbf{+1.922}}
	 & \postext{\textbf{0.269}}
	 & \postext{\textbf{7.144}}
	 & \postext{\textbf{0.0000}}
	 & \postext{\textbf{\mystar\mystar\mystar\mystar}}
	
\\

	 & \blutext{\lrfeature{\textbf{bias}}}
	 & \blutext{\textbf{-1.024}}
	 & \blutext{\textbf{0.291}}
	 & \blutext{\textbf{3.516}}
	 & \blutext{\textbf{0.0004}}
	 & \blutext{\textbf{\mystar\mystar\mystar\mystar}}
	
\\

	 & \negtext{\lrfeature{o\_film/tv}}
	 & \negtext{-0.910}
	 & \negtext{0.470}
	 & \negtext{1.934}
	 & \negtext{0.0531}
	 & \negtext{\textbf{\mystar}}
	
\\

	 & \negtext{\lrfeature{\textbf{multi\_entities}}}
	 & \negtext{\textbf{-0.870}}
	 & \negtext{\textbf{0.165}}
	 & \negtext{\textbf{5.287}}
	 & \negtext{\textbf{0.0000}}
	 & \negtext{\textbf{\mystar\mystar\mystar\mystar}}
	
\\

	 & \negtext{\lrfeature{\textbf{o\_science/tech}}}
	 & \negtext{\textbf{-0.667}}
	 & \negtext{\textbf{0.265}}
	 & \negtext{\textbf{2.522}}
	 & \negtext{\textbf{0.0117}}
	 & \negtext{\textbf{\mystar\mystar}}
	
\\

	 & \negtext{\lrfeature{o\_religion}}
	 & \negtext{-0.655}
	 & \negtext{0.362}
	 & \negtext{1.812}
	 & \negtext{0.0700}
	 & \negtext{\textbf{\mystar}}
	
\\

	 & \postext{\lrfeature{\textbf{o\_writing}}}
	 & \postext{\textbf{+0.402}}
	 & \postext{\textbf{0.189}}
	 & \postext{\textbf{2.128}}
	 & \postext{\textbf{0.0334}}
	 & \postext{\textbf{\mystar\mystar}}
	
\\

	 & \postext{\lrfeature{\textbf{t\_who}}}
	 & \postext{\textbf{+0.363}}
	 & \postext{\textbf{0.183}}
	 & \postext{\textbf{1.990}}
	 & \postext{\textbf{0.0466}}
	 & \postext{\textbf{\mystar\mystar}}
	
\\

\midrule
\multirow{7}{*}{\parbox{2.0 cm}{\textbf{\triviaqa{} \\  \\  \\ Model~Fit: \\ 60.18\%}}}
	 & \blutext{\lrfeature{\textbf{bias}}}
	 & \blutext{\textbf{-1.066}}
	 & \blutext{\textbf{0.114}}
	 & \blutext{\textbf{9.353}}
	 & \blutext{\textbf{0.0000}}
	 & \blutext{\textbf{\mystar\mystar\mystar\mystar}}
	
\\

	 & \negtext{\lrfeature{o\_religion}}
	 & \negtext{-0.443}
	 & \negtext{0.255}
	 & \negtext{1.738}
	 & \negtext{0.0822}
	 & \negtext{\textbf{\mystar}}
	
\\

	 & \postext{\lrfeature{o\_law/crime}}
	 & \postext{+0.412}
	 & \postext{0.218}
	 & \postext{1.890}
	 & \postext{0.0588}
	 & \postext{\textbf{\mystar}}
	
\\

	 & \postext{\lrfeature{\textbf{multi\_answers}}}
	 & \postext{\textbf{+0.341}}
	 & \postext{\textbf{0.024}}
	 & \postext{\textbf{14.090}}
	 & \postext{\textbf{0.0000}}
	 & \postext{\textbf{\mystar\mystar\mystar\mystar}}
	
\\

	 & \postext{\lrfeature{t\_who}}
	 & \postext{+0.230}
	 & \postext{0.129}
	 & \postext{1.778}
	 & \postext{0.0754}
	 & \postext{\textbf{\mystar}}
	
\\

	 & \negtext{\lrfeature{\textbf{o\_politics}}}
	 & \negtext{\textbf{-0.208}}
	 & \negtext{\textbf{0.095}}
	 & \negtext{\textbf{2.177}}
	 & \negtext{\textbf{0.0295}}
	 & \negtext{\textbf{\mystar\mystar}}
	
\\

	 & \postext{\lrfeature{o\_writing}}
	 & \postext{+0.192}
	 & \postext{0.098}
	 & \postext{1.955}
	 & \postext{0.0506}
	 & \postext{\textbf{\mystar}}
	
\\

\bottomrule
        \end{tabular}
    }
    
\caption{\label{tab:logistic-regression} Influential features after filtering \democol{}s based on a $\chi^2$ test (Figure~\ref{fig:accuracies}). Highly influential features ($p$-value < $0.1$), both positive (blue) and negative (red). Higher number of \mystar's signals higher significance.}
\end{table*}

But that is not the only reason an answer may be difficult or easy.  
Following \citet{sugawara-18}, we incorporate features that reveal the questions' difficulty.
For instance, questions that clearly hint  the answer type reduce ambiguity. 
The \hfeat{t\_who} checks if the token ``who'' is in the start of the question.
%
Similarly, \hfeat{t\_what}, \hfeat{t\_when}, and \hfeat{t\_where} capture other entity-types. 
%
%
Questions are also easier if evidence only differs from the question by a couple of words; thus, \hfeat{q\_sim} is the Jaccard similarity between question and evidence tokens. 
Finally, the binary feature \hfeat{e\_train\_count} marks if the person-entities occur in training data more than twice.

We first drop features with negligible effect on accuracy using \lasso{} (regularization $\lambda=1$) by removing  zero coefficients.
For remaining features, Wald statistics~\cite{fahrmeir2007regression} estimate $p$-values. 
Although we initially use quadratic features they are all eliminated during feature reduction.
Thus, we only report the linear features with a minimal significance ($p$-value < 0.1).
%
%

\subsection{How do Properties Affect Accuracy?}

Recall that logistic regression uses features to predict whether the \abr{qa} system will get the answer right or not.
Features associated with correct answers have positive weights (like those derived from \citet{sugawara-18}, \hfeat{q\_sim} and \hfeat{e\_train\_count}), those associated with incorrect answers have negative weights, and features without effect will be near zero.
Among the \hfeat{t\_wh*} features, \hfeat{t\_who} significantly correlates with model correctness, especially in \nq{} and \qb{}, where questions asked directly about a person. 

However, our goal is to see if, \emph{after} accounting for obvious reasons a question could be easy, demographic properties can explain \abr{qa} accuracy. 
The strongest effect is for professions (Table~\ref{tab:logistic-regression}).
For instance, while \nq{} and \qb{} systems struggle on science questions, \triviaqa{}'s does not.
Science has roughly equivalent representation (Table~\ref{tab:demographics}), suggesting \qb{} questions are harder.

While \hfeat{multi\_answer} (and \hfeat{multi\_entities}) reveal harder \nq{} questions, it has a positive effect in \triviaqa{}, as \triviaqa{} uses multiple answers for alternate formulations of answers (Appendix~\ref{appendix:nq-examples}, \ref{appendix:trivia-qa-examples}), which aids machine reading, while multiple \abr{nq} answers are often a sign of ambiguity~\cite{Boyd-Graber-20, Si:Zhao:Boyd-Graber-2021}:
``\emph{Who says that which we call a rose?}'' A: \entity{Juliet}, A: \entity{William Shakespeare}.
For male and female genders, \nq{} has no statistically significant effect on accuracy, only questions about entities with multiple genders depresses accuracy.
%
%
Given the many findings of gender bias in \abr{nlu}~\cite{zhao-17,webster-18,zhao-18,stanovsky-19}, this is surprising.
However, we caution against accepting this conclusion without further investigation given the strong correlation of gender with professional field~\cite{goulden-11}, where we do see significant effects. 

Taken together, the $\chi^{2}$ and logistic regression analysis give us reason to be optimistic: 
although data are skewed for all \demosubset{}s, \abr{qa} systems might well generalize from limited training data across gender and nationality.
%

\section{Related Work}
\label{sec:related}

Language is a reflection of culture.
Like other cultural artifacts---encyclopedias~\cite{reagle-11}, and films~\cite{sap-etal-2017-connotation}---\abr{qa} has more men than women.
Other artifacts like children's books have more gender balance but reflect other aspects of culture~\cite{larrick-65}.

The \abr{nlp} literature is also grappling with demographic discrepancies.
Standard coreference systems falter on gender-balanced corpora~\cite{webster-18}, and \citet{zhao-18} create synthetic training data to reduce bias.
Similar coreference issues plague machine translation systems~\cite{stanovsky-19}, and \citet{li-20} use \abr{qa} to probe biases of \abr{nlp} systems.
\citet{Sen_2020} show that there are shortcomings in \qa{} datasets and evaluations by analysing their out-of-domain generalization capabilities and ability to handle question variation.
Joint models of vision and language suggest that biases come from language, rather than from  vision~\cite{ross-etal-2021-measuring}.
However, despite a range of mitigation techniques~\citep[inter alia]{zhao-17} none, to our knowledge, have been successfully applied to \abr{qa}, especially from the demographic viewpoint.

\section{Discussion and Conclusion}
\label{sec:conclusion}

This paper delivers both good news and bad news.
While datasets remain imperfect and reflect societal imperfections, for many demographic properties, we do not find strong evidence that \abr{qa} suffers from this skew.

However, this is an absence of evidence rather than evidence of absence: these are skewed datasets that have fewer than a quarter of the questions about women.
It is difficult to make confident assessments on such small datasets---many demographic \demorow{}s were excluded because they appeared infrequently (or not at all).
Improving the diversity of \abr{qa} datasets can help us be more certain that \abr{qa} systems do generalize and reflect the diverse human experience.
Considering such shortcomings, \citet{Rodriguez_2021} advocate improving evaluation by focusing on more important examples for ranking models; demographic properties could further refine more holistic evaluations.

A broader analysis beyond person entities would indeed be a natural extension of this work. Label propagation can expand the analysis beyond people: the \entity{Hershey-Chase} experiment is associated with \entity{Alfred Hershey} and \entity{Martha Chase}, so it would—given the neighboring entities in the Wikipedia link graph—be 100\% American, 50\% male, and 50\% female.
Another direction for future work is accuracy under counterfactual perturbation: swapping real-world entities (in contrast with nonce entities in \citet{li-20}) with different demographic values.

%
Nonetheless, particularly for professional fields, imbalances remain.
The lack of representation in \abr{qa} could cause us to think that things are better than they are because of Simpson's paradox~\cite{blyth-72}: gender and profession are not independent!
For example, in \nq{}, our accuracy on women is higher in part because of its tilt toward entertainment, and we cannot say much about women scientists.
We therefore caution against interpreting strong model performance on existing \abr{qa} datasets as evidence that the task is ‘solved’.
Instead, future work must consider better dataset construction strategies and robustness of accuracy metrics to different \demosubset{}s of available data, as well as unseen examples.

%
%
%
%
%
%
\section*{Acknowledgements}
 We thank Michael Collins, Slav Petrov, Tulsee Doshi, Sephora Madjiheurem, Benjamin B\"orschinger, Pedro Rodriguez, Massimiliano Ciaramita, Kenton Lee, Alex Beutal, Kenton Lee, and Emily Pitler for their early and insightful comments on the proposal and drafts. Additionally, insights about Google's \abr{Cloud-NL} entity linking tool and \abr{WikiData KB} from Jan Botha, Denny Vrandecic, Livio Soares and Tom Kwiatkowski were quite useful in designing the entity linking and attribute extraction pipeline. 
\section*{Ethical Considerations}
\label{sec:ethics}

This work analyses demographic subsets across \qa{} datasets based on Gender, Nationality and Profession. 
We believe the work makes a positive contribution to representation and diversity by pointing out the skewed distribution of existing \abr{qa} datasets.
To avoid noise being interpreted as signal given the lack of diversity in these datasets, we could not include various subgroups that we believe should have been part of this study: non-binary, intersectional groups (e.g., women scientists in \abr{nq}), people indigenous to subnational regions, etc. 
We believe increasing representation of all such groups in \abr{qa} datasets would improve upon the status quo.
We infer properties of mentions using Google Cloud-\abr{nl} to link the entity in a \abr{qa} example to an entry in the \textsc{WikiData} knowledge base to attribute gender, profession and nationality. 
We acknowledge that this is not foolproof and itself vulnerable to bias, although our small-scale accuracy evaluation did not reveal any concerning patterns.

All human annotations are provided by authors to verify entity-linkings and were fairly compensated.


\bibliography{bib/journal-full,bib/jbg,bib/maharshi}

\begin{thebibliography}{38}
\expandafter\ifx\csname natexlab\endcsname\relax\def\natexlab#1{#1}\fi

\bibitem[{Alberti et~al.(2019)Alberti, Lee, and Collins}]{alberti2019bert}
Chris Alberti, Kenton Lee, and Michael Collins. 2019.
\newblock A {BERT} baseline for the natural questions.
\newblock \emph{arXiv preprint arXiv:1901.08634}.

\bibitem[{Artetxe et~al.(2019)Artetxe, Ruder, and Yogatama}]{artetxe2019xquad}
Mikel Artetxe, Sebastian Ruder, and Dani Yogatama. 2019.
\newblock On the cross-lingual transferability of monolingual representations.
\newblock \emph{CoRR}, abs/1910.11856.

\bibitem[{Blyth(1972)}]{blyth-72}
Colin~R. Blyth. 1972.
\newblock \href {http://www.jstor.org/stable/2284382} {On {S}impson's paradox
  and the sure-thing principle}.
\newblock \emph{Journal of the American Statistical Association},
  67(338):364--366.

\bibitem[{Boyd-Graber and B\"orschinger(2020)}]{Boyd-Graber-20}
Jordan Boyd-Graber and Benjamin B\"orschinger. 2020.
\newblock \href {http://umiacs.umd.edu/~jbg//docs/2020_acl_trivia.pdf} {What
  question answering can learn from trivia nerds}.
\newblock In \emph{Proceedings of the Association for Computational
  Linguistics}.

\bibitem[{Boyd-Graber et~al.(2012)Boyd-Graber, Satinoff, He, and
  III}]{boyd-graber-12}
Jordan Boyd-Graber, Brianna Satinoff, He~He, and Hal~Daume III. 2012.
\newblock Besting the quiz master: Crowdsourcing incremental classification
  games.
\newblock In \emph{Proceedings of Empirical Methods in Natural Language
  Processing}.

\bibitem[{Fahrmeir et~al.(2007)Fahrmeir, Kneib, Lang, and
  Marx}]{fahrmeir2007regression}
Ludwig Fahrmeir, Thomas Kneib, Stefan Lang, and Brian Marx. 2007.
\newblock \emph{Regression}.
\newblock Springer.

\bibitem[{Ferrucci et~al.(2010)Ferrucci, Brown, Chu-Carroll, Fan, Gondek,
  Kalyanpur, Lally, Murdock, Nyberg, Prager, Schlaefer, and Welty}]{ferruci-10}
David Ferrucci, Eric Brown, Jennifer Chu-Carroll, James Fan, David Gondek,
  Aditya~A. Kalyanpur, Adam Lally, J.~William Murdock, Eric Nyberg, John
  Prager, Nico Schlaefer, and Chris Welty. 2010.
\newblock {Building Watson: An Overview of the DeepQA Project}.
\newblock \emph{AI Magazine}, 31(3).

\bibitem[{Glickman and Jones(1999)}]{glickman-99}
Mark~E Glickman and Albyn~C Jones. 1999.
\newblock \href
  {http://scholar.google.de/scholar.bib?q=info:r4leZxFtllwJ:scholar.google.com/&output=citation&scisig=AAGBfm0AAAAAUm2XckUml2ofzJwFf4A_kI-bLVVMwfdI&scisf=4&hl=en&scfhb=1}
  {Rating the chess rating system}.
\newblock \emph{Chance}, 12.

\bibitem[{Goulden et~al.(2011)Goulden, Mason, and Frasch}]{goulden-11}
Marc Goulden, Mary~Ann Mason, and Karie Frasch. 2011.
\newblock \href {http://www.jstor.org/stable/41328583} {Keeping women in the
  science pipeline}.
\newblock \emph{The Annals of the American Academy of Political and Social
  Science}, 638:141--162.

\bibitem[{Holm(1979)}]{holm-79}
Sture Holm. 1979.
\newblock \href {http://www.jstor.org/stable/4615733} {A simple sequentially
  rejective multiple test procedure}.
\newblock \emph{Scandinavian Journal of Statistics}, 6(2):65--70.

\bibitem[{Joshi et~al.(2017)Joshi, Choi, Weld, and Zettlemoyer}]{joshi-17}
Mandar Joshi, Eunsol Choi, Daniel Weld, and Luke Zettlemoyer. 2017.
\newblock \href {https://www.aclweb.org/anthology/P17-1147} {{TriviaQA}: A
  large scale distantly supervised challenge dataset for reading
  comprehension}.
\newblock In \emph{Proceedings of the Association for Computational
  Linguistics}.

\bibitem[{Kiss et~al.(2010)Kiss, Ke{\ss}elmeier, M{\"u}ller, Roch, Stadtfeld,
  and Strunk}]{kiss-10}
Tibor Kiss, Katja Ke{\ss}elmeier, Antje M{\"u}ller, Claudia Roch, Tobias
  Stadtfeld, and Jan Strunk. 2010.
\newblock \href {https://www.aclweb.org/anthology/C10-2064} {A logistic
  regression model of determiner omission in {PP}s}.
\newblock In \emph{Coling 2010: Posters}, pages 561--569, Beijing, China.
  Coling 2010 Organizing Committee.

\bibitem[{Kwiatkowski et~al.(2019)Kwiatkowski, Palomaki, Redfield, Collins,
  Parikh, Alberti, Epstein, Polosukhin, Kelcey, Devlin, Lee, Toutanova, Jones,
  Chang, Dai, Uszkoreit, Le, and Petrov}]{kwiatkowski-19}
Tom Kwiatkowski, Jennimaria Palomaki, Olivia Redfield, Michael Collins, Ankur
  Parikh, Chris Alberti, Danielle Epstein, Illia Polosukhin, Matthew Kelcey,
  Jacob Devlin, Kenton Lee, Kristina~N. Toutanova, Llion Jones, Ming-Wei Chang,
  Andrew Dai, Jakob Uszkoreit, Quoc Le, and Slav Petrov. 2019.
\newblock \href
  {https://tomkwiat.users.x20web.corp.google.com/papers/natural-questions/main-1455-kwiatkowski.pdf}
  {Natural questions: {A} benchmark for question answering research}.
\newblock \emph{Transactions of the Association for Computational Linguistics}.

\bibitem[{Larrick(1965)}]{larrick-65}
Nancy Larrick. 1965.
\newblock The all-white world of children's books.
\newblock \emph{The Saturday Review}, 64:63--65.

\bibitem[{Lee et~al.(2019)Lee, Chang, and Toutanova}]{lee2019latent}
Kenton Lee, Ming-Wei Chang, and Kristina Toutanova. 2019.
\newblock \href {https://doi.org/10.18653/v1/P19-1612} {Latent retrieval for
  weakly supervised open domain question answering}.
\newblock In \emph{Proceedings of the Association for Computational
  Linguistics}.

\bibitem[{Lewis et~al.(2020)Lewis, Oguz, Rinott, Riedel, and
  Schwenk}]{lewis-etal-2020-mlqa}
Patrick Lewis, Barlas Oguz, Ruty Rinott, Sebastian Riedel, and Holger Schwenk.
  2020.
\newblock \href {https://doi.org/10.18653/v1/2020.acl-main.653} {{MLQA}:
  Evaluating cross-lingual extractive question answering}.
\newblock In \emph{Proceedings of the 58th Annual Meeting of the Association
  for Computational Linguistics}, pages 7315--7330, Online. Association for
  Computational Linguistics.

\bibitem[{Li et~al.(2020)Li, Khashabi, Khot, Sabharwal, and Srikumar}]{li-20}
Tao Li, Daniel Khashabi, Tushar Khot, Ashish Sabharwal, and Vivek Srikumar.
  2020.
\newblock \href {https://doi.org/10.18653/v1/2020.findings-emnlp.311}
  {{UNQOVER}ing stereotyping biases via underspecified questions}.
\newblock In \emph{Findings of the Association for Computational Linguistics:
  EMNLP}.

\bibitem[{McCloskey and McCloskey(1987)}]{mccloskey-87}
Deirdre~N. McCloskey and Donald~N. McCloskey. 1987.
\newblock \href {https://books.google.com/books?id=7g-xAAAAIAAJ}
  {\emph{Econometric History}}.
\newblock Casebook Series. Macmillan Education.

\bibitem[{Min et~al.(2020)Min, Michael, Hajishirzi, and
  Zettlemoyer}]{min-etal-2020-ambigqa}
Sewon Min, Julian Michael, Hannaneh Hajishirzi, and Luke Zettlemoyer. 2020.
\newblock \href {https://doi.org/10.18653/v1/2020.emnlp-main.466} {{A}mbig{QA}:
  Answering ambiguous open-domain questions}.
\newblock In \emph{Proceedings of the 2020 Conference on Empirical Methods in
  Natural Language Processing (EMNLP)}, pages 5783--5797, Online. Association
  for Computational Linguistics.

\bibitem[{Najberg(2018)}]{najberg-18}
Adam Najberg. 2018.
\newblock \href
  {https://www.alizila.com/alibaba-ai-model-tops-humans-in-reading-comprehension/}
  {{Alibaba} {AI} model tops humans in reading comprehension}.

\bibitem[{Niculescu-Mizil and Caruana(2005)}]{niculescu2005predicting}
Alexandru Niculescu-Mizil and Rich Caruana. 2005.
\newblock Predicting good probabilities with supervised learning.
\newblock In \emph{Proceedings of the International Conference of Machine
  Learning}.

\bibitem[{Plackett(1983)}]{plackett1983karl}
Robin~L Plackett. 1983.
\newblock Karl {P}earson and the chi-squared test.
\newblock \emph{International Statistical Review/Revue Internationale de
  Statistique}, pages 59--72.

\bibitem[{Poole and Rosenthal(2011)}]{poole-11}
K.T. Poole and H.L. Rosenthal. 2011.
\newblock \href {https://books.google.com/books?id=eOuZqa7UbqUC}
  {\emph{Ideology and Congress}}.
\newblock American Studies. Transaction Publishers.

\bibitem[{Rajpurkar et~al.(2016)Rajpurkar, Zhang, Lopyrev, and
  Liang}]{rajpurkar-16}
Pranav Rajpurkar, Jian Zhang, Konstantin Lopyrev, and Percy Liang. 2016.
\newblock {SQuAD}: 100,000+ questions for machine comprehension of text.
\newblock In \emph{Proceedings of Empirical Methods in Natural Language
  Processing}.

\bibitem[{Reagle and Rhue(2011)}]{reagle-11}
Joseph Reagle and Lauren Rhue. 2011.
\newblock \href {https://ijoc.org/index.php/ijoc/article/view/777} {Gender bias
  in {Wikipedia} and {Britannica}}.
\newblock \emph{International Journal of Communication}, 5(0).

\bibitem[{Rodriguez et~al.(2021)Rodriguez, Barrow, Hoyle, Lalor, Jia, and
  Boyd-Graber}]{Rodriguez_2021}
Pedro Rodriguez, Joe Barrow, Alexander~Miserlis Hoyle, John~P. Lalor, Robin
  Jia, and Jordan Boyd-Graber. 2021.
\newblock \href {https://doi.org/10.18653/v1/2021.acl-long.346} {Evaluation
  examples are not equally informative: How should that change nlp
  leaderboards?}
\newblock \emph{Proceedings of the 59th Annual Meeting of the Association for
  Computational Linguistics and the 11th International Joint Conference on
  Natural Language Processing (Volume 1: Long Papers)}.

\bibitem[{Ross et~al.(2021)Ross, Katz, and Barbu}]{ross-etal-2021-measuring}
Candace Ross, Boris Katz, and Andrei Barbu. 2021.
\newblock \href {https://doi.org/10.18653/v1/2021.naacl-main.78} {Measuring
  social biases in grounded vision and language embeddings}.
\newblock In \emph{Proceedings of the 2021 Conference of the North American
  Chapter of the Association for Computational Linguistics: Human Language
  Technologies}, pages 998--1008, Online. Association for Computational
  Linguistics.

\bibitem[{Sap et~al.(2017)Sap, Prasettio, Holtzman, Rashkin, and
  Choi}]{sap-etal-2017-connotation}
Maarten Sap, Marcella~Cindy Prasettio, Ari Holtzman, Hannah Rashkin, and Yejin
  Choi. 2017.
\newblock \href {https://doi.org/10.18653/v1/D17-1247} {Connotation frames of
  power and agency in modern films}.
\newblock In \emph{Proceedings of Empirical Methods in Natural Language
  Processing}.

\bibitem[{Sen and Saffari(2020)}]{Sen_2020}
Priyanka Sen and Amir Saffari. 2020.
\newblock \href {https://doi.org/10.18653/v1/2020.emnlp-main.190} {What do
  models learn from question answering datasets?}
\newblock \emph{Proceedings of the 2020 Conference on Empirical Methods in
  Natural Language Processing (EMNLP)}.

\bibitem[{Si et~al.(2021)Si, Zhao, and Boyd-Graber}]{Si:Zhao:Boyd-Graber-2021}
Chenglei Si, Chen Zhao, and Jordan Boyd-Graber. 2021.
\newblock What's in a name? answer equivalence for open-domain question
  answering.
\newblock In \emph{Emperical Methods in Natural Language Processing}.

\bibitem[{Stanovsky et~al.(2019)Stanovsky, Smith, and
  Zettlemoyer}]{stanovsky-19}
Gabriel Stanovsky, Noah~A. Smith, and Luke Zettlemoyer. 2019.
\newblock \href {https://doi.org/10.18653/v1/P19-1164} {Evaluating gender bias
  in machine translation}.
\newblock In \emph{Proceedings of the Association for Computational
  Linguistics}.

\bibitem[{Sugawara et~al.(2018)Sugawara, Inui, Sekine, and
  Aizawa}]{sugawara-18}
Saku Sugawara, Kentaro Inui, Satoshi Sekine, and Akiko Aizawa. 2018.
\newblock \href {https://doi.org/10.18653/v1/D18-1453} {What makes reading
  comprehension questions easier?}
\newblock In \emph{Proceedings of Empirical Methods in Natural Language
  Processing}.

\bibitem[{van~der Linden and Hambleton(2013)}]{Linden-2013}
Wim~J van~der Linden and Ronald~K Hambleton. 2013.
\newblock \emph{Handbook of modern item response theory}.
\newblock Springer Science \& Business Media.

\bibitem[{Wallace et~al.(2019)Wallace, Rodriguez, Feng, Yamada, and
  Boyd-Graber}]{wallace-19}
Eric Wallace, Pedro Rodriguez, Shi Feng, Ikuya Yamada, and Jordan Boyd-Graber.
  2019.
\newblock Trick me if you can: Human-in-the-loop generation of adversarial
  question answering examples.
\newblock \emph{Transactions of the Association of Computational Linguistics},
  10.

\bibitem[{Webster et~al.(2018)Webster, Recasens, Axelrod, and
  Baldridge}]{webster-18}
Kellie Webster, Marta Recasens, Vera Axelrod, and Jason Baldridge. 2018.
\newblock \href {https://doi.org/10.1162/tacl_a_00240} {Mind the {GAP}: A
  balanced corpus of gendered ambiguous pronouns}.
\newblock \emph{Transactions of the Association for Computational Linguistics},
  6:605--617.

\bibitem[{Yogatama et~al.(2011)Yogatama, Heilman, O{'}Connor, Dyer, Routledge,
  and Smith}]{yogatama-11}
Dani Yogatama, Michael Heilman, Brendan O{'}Connor, Chris Dyer, Bryan~R.
  Routledge, and Noah~A. Smith. 2011.
\newblock \href {https://www.aclweb.org/anthology/D11-1055} {Predicting a
  scientific community{'}s response to an article}.
\newblock In \emph{Proceedings of Empirical Methods in Natural Language
  Processing}.

\bibitem[{Zhao et~al.(2017)Zhao, Wang, Yatskar, Ordonez, and Chang}]{zhao-17}
Jieyu Zhao, Tianlu Wang, Mark Yatskar, Vicente Ordonez, and Kai-Wei Chang.
  2017.
\newblock \href {https://www.aclweb.org/anthology/D17-1323} {Men also like
  shopping: Reducing gender bias amplification using corpus-level constraints}.
\newblock In \emph{Proceedings of Empirical Methods in Natural Language
  Processing}.

\bibitem[{Zhao et~al.(2018)Zhao, Wang, Yatskar, Ordonez, and Chang}]{zhao-18}
Jieyu Zhao, Tianlu Wang, Mark Yatskar, Vicente Ordonez, and Kai-Wei Chang.
  2018.
\newblock \href {https://doi.org/10.18653/v1/N18-2003} {Gender bias in
  coreference resolution: Evaluation and debiasing methods}.
\newblock In \emph{Conference of the North American Chapter of the Association
  for Computational Linguistics}.

\end{thebibliography}
\bibliographystyle{style/acl_natbib}

\clearpage
\appendix\section*{Appendix}

\section{Entity collapses of demographic \demorow{}s}
While mapping \qa{} examples to person entities and values for their corresponding demographic characteristics (Section \ref{sec:mapping}), we encountered many nearby values: `Poet', `Writer', `Author'. We collapse such values into a single label which we use for further analysis. This section enlists all the collapses that we encounter for determining nationality of people (Appendix~\ref{appendix:country-collapse}) and their professions (Appendix~\ref{appendix:professions-collapse}).

\subsection{Entity-collapses for Nationality values}
\label{appendix:country-collapse}

\paragraph{\textbf{\texttt{US:}}}
\feature{kingdom of hawaii}, \feature{united states}, \feature{united states of america}\\
\paragraph{\textbf{\texttt{UK:}}}
\feature{commonwealth of england}, \feature{great britain}, \feature{kingdom of england}, \feature{kingdom of mercia}, \feature{kingdom of scotland}, \feature{kingdom of wessex}, \feature{united kingdom}, \feature{united kingdom of great britain and ireland}\\
\paragraph{\textbf{\texttt{Albania:}}}
\feature{kingdom of albania}\\
\paragraph{\textbf{\texttt{Austria:}}}
\feature{austrian empire}, \feature{federal state of austria}, \feature{first republic of austria}\\
\paragraph{\textbf{\texttt{Cyprus:}}}
\feature{kingdom of cyprus}, \feature{republic of cyprus}, \feature{turkish republic of northern cyprus}\\
\paragraph{\textbf{\texttt{Denmark:}}}
\feature{kingdom of denmark}\\
\paragraph{\textbf{\texttt{France:}}}
\feature{kingdom of france}\\
\paragraph{\textbf{\texttt{Germany:}}}
\feature{german confederation}, \feature{german democratic republic}, \feature{german empire}, \feature{german reich}, \feature{germany}, \feature{kingdom of hanover}, \feature{kingdom of prussia}, \feature{kingdom of saxony}, \feature{nazi germany}, \feature{north german confederation}, \feature{prussia}, \feature{republic of german-austria}, \feature{west germany}\\
\paragraph{\textbf{\texttt{Greece:}}}
\feature{ancient greece}, \feature{greece}\\
\paragraph{\textbf{\texttt{Hungary:}}}
\feature{hungary}, \feature{kingdom of hungary}, \feature{people's republic of hungary}\\
\paragraph{\textbf{\texttt{Ireland:}}}
\feature{irish republic}, \feature{kingdom of ireland}\\
\paragraph{\textbf{\texttt{Italy:}}}
\feature{ancient rome}, \feature{florence}, \feature{holy roman empire}, \feature{kingdom of italy}, \feature{kingdom of sardinia}\\
\paragraph{\textbf{\texttt{Netherlands:}}}
\feature{dutch republic}, \feature{kingdom of the netherlands}\\
\paragraph{\textbf{\texttt{Poland:}}}
\feature{kingdom of poland}, \feature{poland}\\
\paragraph{\textbf{\texttt{Portugal:}}}
\feature{kingdom of portugal}\\
\paragraph{\textbf{\texttt{Romania:}}}
\feature{kingdom of romania}, \feature{romania}, \feature{socialist republic of romania}\\
\paragraph{\textbf{\texttt{Spain:}}}
\feature{crown of castile}, \feature{kingdom of aragon}, \feature{kingdom of castile}, \feature{kingdom of navarre}, \feature{spain}\\
\paragraph{\textbf{\texttt{Yugoslavia:}}}
\feature{federal republic of yugoslavia}, \feature{kingdom of yugoslavia}, \feature{socialist federal republic of yugoslavia}, \feature{yugoslavia}\\
\paragraph{\textbf{\texttt{Iraq:}}}
\feature{ba'athist iraq}, \feature{iraq}, \feature{kingdom of iraq}, \feature{mandatory iraq}, \feature{republic of iraq (1958–68)}\\
\paragraph{\textbf{\texttt{Israel:}}}
\feature{israel}, \feature{kingdom of israel}, \feature{land of israel}\\
\paragraph{\textbf{\texttt{Russia:}}}
\feature{russia}, \feature{russian empire}, \feature{russian soviet federative socialist republic}, \feature{soviet union}, \feature{tsardom of russia}\\
\paragraph{\textbf{\texttt{India:}}}
\feature{british raj}, \feature{delhi sultanate}, \feature{dominion of india}, \feature{india}\\
\paragraph{\textbf{\texttt{China:}}}
\feature{china}, \feature{people's republic of china}, \feature{republic of china (1912–1949)}\\
\paragraph{\textbf{\texttt{Egypt:}}}
\feature{ancient egypt}, \feature{egypt}, \feature{kingdom of egypt}, \feature{republic of egypt}\\
\subsection{Entity-collapses for \textit{Profession} values}
\label{appendix:professions-collapse}

\paragraph{\textbf{\texttt{Writing:}}}
\feature{author}, \feature{biographer}, \feature{cartoonist}, \feature{children's writer}, \feature{comedy writer}, \feature{comics artist}, \feature{comics writer}, \feature{contributing editor}, \feature{cookery writer}, \feature{detective writer}, \feature{diarist}, \feature{editor}, \feature{editorial columnist}, \feature{essayist}, \feature{fairy tales writer}, \feature{grammarian}, \feature{hymnwriter}, \feature{journalist}, \feature{lexicographer}, \feature{librettist}, \feature{linguist}, \feature{literary}, \feature{literary critic}, \feature{literary editor}, \feature{literary scholar}, \feature{memoirist}, \feature{newspaper editor}, \feature{non-fiction writer}, \feature{novelist}, \feature{opinion journalist}, \feature{philologist}, \feature{photojournalist}, \feature{physician writer}, \feature{playwright}, \feature{poet}, \feature{poet lawyer}, \feature{preface author}, \feature{prosaist}, \feature{religious writer}, \feature{science fiction writer}, \feature{science writer}, \feature{scientific editor}, \feature{screenwriter}, \feature{short story writer}, \feature{tragedy writer}, \feature{travel writer}, \feature{women letter writer}, \feature{writer}\\
\paragraph{\textbf{\texttt{Sports:}}}
\feature{amateur wrestler}, \feature{american football coach}, \feature{american football player}, \feature{archer}, \feature{artistic gymnast}, \feature{association football manager}, \feature{association football player}, \feature{association football referee}, \feature{athlete}, \feature{athletics competitor}, \feature{australian rules football player}, \feature{badminton player}, \feature{ballet dancer}, \feature{ballet master}, \feature{ballet pedagogue}, \feature{baseball player}, \feature{basketball coach}, \feature{basketball player}, \feature{biathlete}, \feature{biathlon coach}, \feature{boxer}, \feature{bridge player}, \feature{canadian football player}, \feature{chess player}, \feature{choreographer}, \feature{coach}, \feature{cricket umpire}, \feature{cricketer}, \feature{dancer}, \feature{darts player}, \feature{field hockey player}, \feature{figure skater}, \feature{figure skating choreographer}, \feature{figure skating coach}, \feature{formula one driver}, \feature{gaelic football player}, \feature{golfer}, \feature{gridiron football player}, \feature{gymnast}, \feature{head coach}, \feature{hurler}, \feature{ice dancer}, \feature{ice hockey coach}, \feature{ice hockey player}, \feature{jockey}, \feature{judoka}, \feature{lacrosse player}, \feature{long-distance runner}, \feature{marathon runner}, \feature{marimba player}, \feature{martial artist}, \feature{middle-distance runner}, \feature{mixed martial artist}, \feature{motorcycle racer}, \feature{poker player}, \feature{polo player}, \feature{pool player}, \feature{professional wrestler}, \feature{quidditch player}, \feature{racing automobile driver}, \feature{racing driver}, \feature{rink hockey player}, \feature{rugby league player}, \feature{rugby player}, \feature{rugby union coach}, \feature{rugby union player}, \feature{runner}, \feature{short track speed skater}, \feature{skateboarder}, \feature{skeleton racer}, \feature{snooker player}, \feature{snowboarder}, \feature{sport cyclist}, \feature{sport shooter}, \feature{sporting director}, \feature{sports agent}, \feature{sports commentator}, \feature{sprinter}, \feature{squash player}, \feature{surfer}, \feature{swimmer}, \feature{table tennis player}, \feature{taekwondo athlete}, \feature{tennis coach}, \feature{tennis player}, \feature{thai boxer}, \feature{track and field coach}, \feature{viol player}, \feature{volleyball player}, \feature{water polo player}\\
\paragraph{\textbf{\texttt{Music:}}}
\feature{bass guitar}, \feature{bassist}, \feature{blues musician}, \feature{child singer}, \feature{classical composer}, \feature{classical guitarist}, \feature{classical pianist}, \feature{collector of folk music}, \feature{composer}, \feature{conductor}, \feature{country musician}, \feature{drummer}, \feature{film score composer}, \feature{ghost singer}, \feature{guitar maker}, \feature{guitarist}, \feature{heavy metal singer}, \feature{instrument maker}, \feature{instrumentalist}, \feature{jazz guitarist}, \feature{jazz musician}, \feature{jazz singer}, \feature{keyboardist}, \feature{lyricist}, \feature{multi-instrumentalist}, \feature{music arranger}, \feature{music artist}, \feature{music critic}, \feature{music director}, \feature{music interpreter}, \feature{music pedagogue}, \feature{music pedagogy}, \feature{music producer}, \feature{music publisher}, \feature{music theorist}, \feature{music video director}, \feature{musical}, \feature{musical instrument maker}, \feature{musician}, \feature{musicologist}, \feature{opera composer}, \feature{opera singer}, \feature{optical instrument maker}, \feature{organist}, \feature{pianist}, \feature{playback singer}, \feature{professor of music composition}, \feature{rapper}, \feature{record producer}, \feature{recording artist}, \feature{rock drummer}, \feature{rock musician}, \feature{saxophonist}, \feature{session musician}, \feature{singer}, \feature{singer-songwriter}, \feature{songwriter}, \feature{violinist}\\
\paragraph{\textbf{\texttt{Fictional:}}}
\feature{fictional aviator}, \feature{fictional businessperson}, \feature{fictional character}, \feature{fictional cowboy}, \feature{fictional domestic worker}, \feature{fictional firefighter}, \feature{fictional journalist}, \feature{fictional mass murderer}, \feature{fictional pirate}, \feature{fictional police officer}, \feature{fictional politician}, \feature{fictional schoolteacher}, \feature{fictional scientist}, \feature{fictional seaman}, \feature{fictional secretary}, \feature{fictional soldier}, \feature{fictional space traveller}, \feature{fictional taxi driver}, \feature{fictional vigilante}, \feature{fictional waitperson}, \feature{fictional writer}\\
\paragraph{\textbf{\texttt{Politics:}}}
\feature{activist}, \feature{ambassador}, \feature{animal rights advocate}, \feature{anti-vaccine activist}, \feature{civil rights advocate}, \feature{civil servant}, \feature{climate activist}, \feature{colonial administrator}, \feature{consort}, \feature{dictator}, \feature{diplomat}, \feature{drag queen}, \feature{duke}, \feature{emperor}, \feature{feminist}, \feature{foreign minister}, \feature{government agent}, \feature{governor}, \feature{human rights activist}, \feature{internet activist}, \feature{khan}, \feature{king}, \feature{leader}, \feature{lgbt rights activist}, \feature{military commander}, \feature{military leader}, \feature{military officer}, \feature{military personnel}, \feature{military theorist}, \feature{minister}, \feature{monarch}, \feature{peace activist}, \feature{political activist}, \feature{political philosopher}, \feature{political scientist}, \feature{political theorist}, \feature{politician}, \feature{president}, \feature{prince}, \feature{princess}, \feature{protestant reformer}, \feature{queen}, \feature{queen consort}, \feature{queen regnant}, \feature{religious leader}, \feature{revolutionary}, \feature{ruler}, \feature{secretary}, \feature{social reformer}, \feature{socialite}, \feature{tribal chief}\\
\paragraph{\textbf{\texttt{Artist:}}}
\feature{architect}, \feature{artist}, \feature{baker}, \feature{blacksmith}, \feature{car designer}, \feature{chef}, \feature{costume designer}, \feature{design}, \feature{designer}, \feature{fashion designer}, \feature{fashion photographer}, \feature{fresco painter}, \feature{furniture designer}, \feature{game designer}, \feature{glass artist}, \feature{goldsmith}, \feature{graffiti artist}, \feature{graphic artist}, \feature{graphic designer}, \feature{house painter}, \feature{illustrator}, \feature{industrial designer}, \feature{interior designer}, \feature{jewellery designer}, \feature{landscape architect}, \feature{landscape painter}, \feature{lighting designer}, \feature{painter}, \feature{photographer}, \feature{postage stamp designer}, \feature{printmaker}, \feature{production designer}, \feature{scientific illustrator}, \feature{sculptor}, \feature{sound designer}, \feature{textile designer}, \feature{type designer}, \feature{typographer}, \feature{visual artist}\\
\paragraph{\textbf{\texttt{Film/tv:}}}
\feature{actor}, \feature{character actor}, \feature{child actor}, \feature{documentary filmmaker}, \feature{dub actor}, \feature{factory owner}, \feature{fashion model}, \feature{film actor}, \feature{film critic}, \feature{film director}, \feature{film editor}, \feature{film producer}, \feature{filmmaker}, \feature{glamour model}, \feature{line producer}, \feature{model}, \feature{pornographic actor}, \feature{reality television participant}, \feature{runway model}, \feature{television actor}, \feature{television director}, \feature{television editor}, \feature{television presenter}, \feature{television producer}, \feature{voice actor}\\
\paragraph{\textbf{\texttt{Executive:}}}
\feature{bank manager}, \feature{business executive}, \feature{business magnate}, \feature{businessperson}, \feature{chief executive officer}, \feature{entrepreneur}, \feature{executive officer}, \feature{executive producer}, \feature{manager}, \feature{real estate entrepreneur}, \feature{talent manager}\\
\paragraph{\textbf{\texttt{Stage:}}}
\feature{circus performer}, \feature{comedian}, \feature{entertainer}, \feature{mime artist}, \feature{musical theatre actor}, \feature{stage actor}, \feature{stand-up comedian}, \feature{theater director}\\
\paragraph{\textbf{\texttt{Law/crime:}}}
\feature{art thief}, \feature{attorney at law}, \feature{bank robber}, \feature{canon law jurist}, \feature{courtier}, \feature{criminal}, \feature{judge}, \feature{jurist}, \feature{lawyer}, \feature{official}, \feature{private investigator}, \feature{robber}, \feature{serial killer}, \feature{spy}, \feature{thief}, \feature{war criminal}\\
\paragraph{\textbf{\texttt{History:}}}
\feature{anthropologist}, \feature{archaeologist}, \feature{art historian}, \feature{church historian}, \feature{classical archaeologist}, \feature{egyptologist}, \feature{explorer}, \feature{historian}, \feature{historian of classical antiquity}, \feature{historian of mathematics}, \feature{historian of science}, \feature{historian of the modern age}, \feature{labor historian}, \feature{legal historian}, \feature{literary historian}, \feature{military historian}, \feature{music historian}, \feature{paleoanthropologist}, \feature{paleontologist}, \feature{philosophy historian}, \feature{polar explorer}, \feature{scientific explorer}\\
\paragraph{\textbf{\texttt{Science/tech:}}}
\feature{aerospace engineer}, \feature{alchemist}, \feature{anesthesiologist}, \feature{artificial intelligence researcher}, \feature{astrologer}, \feature{astronaut}, \feature{astronomer}, \feature{astrophysicist}, \feature{auto mechanic}, \feature{bacteriologist}, \feature{biochemist}, \feature{biologist}, \feature{botanist}, \feature{bryologist}, \feature{cardiologist}, \feature{chemical engineer}, \feature{chemist}, \feature{chief engineer}, \feature{civil engineer}, \feature{climatologist}, \feature{cognitive scientist}, \feature{combat engineer}, \feature{computer scientist}, \feature{cosmologist}, \feature{crystallographer}, \feature{earth scientist}, \feature{ecologist}, \feature{educational psychologist}, \feature{electrical engineer}, \feature{engineer}, \feature{environmental scientist}, \feature{epidemiologist}, \feature{ethnologist}, \feature{ethologist}, \feature{evolutionary biologist}, \feature{geochemist}, \feature{geographer}, \feature{geologist}, \feature{geophysicist}, \feature{immunologist}, \feature{industrial engineer}, \feature{inventor}, \feature{marine biologist}, \feature{mathematician}, \feature{mechanic}, \feature{mechanical automaton engineer}, \feature{mechanical engineer}, \feature{meteorologist}, \feature{microbiologist}, \feature{mining engineer}, \feature{naturalist}, \feature{neurologist}, \feature{neuroscientist}, \feature{nuclear physicist}, \feature{nurse}, \feature{ontologist}, \feature{ornithologist}, \feature{patent inventor}, \feature{pharmacologist}, \feature{physician}, \feature{physicist}, \feature{physiologist}, \feature{planetary scientist}, \feature{psychiatrist}, \feature{psychoanalyst}, \feature{psychologist}, \feature{railroad engineer}, \feature{railway engineer}, \feature{research assistant}, \feature{researcher}, \feature{scientist}, \feature{social psychologist}, \feature{social scientist}, \feature{sociologist}, \feature{software engineer}, \feature{space scientist}, \feature{statistician}, \feature{structural engineer}, \feature{theoretical biologist}, \feature{theoretical physicist}, \feature{virologist}, \feature{zoologist}\\
\paragraph{\textbf{\texttt{Education:}}}
\feature{academic}, \feature{adjunct professor}, \feature{associate professor}, \feature{educator}, \feature{head teacher}, \feature{high school teacher}, \feature{history teacher}, \feature{lady margaret's professor of divinity}, \feature{pedagogue}, \feature{professor}, \feature{school teacher}, \feature{sex educator}, \feature{teacher}, \feature{university teacher}\\
\paragraph{\textbf{\texttt{Economics:}}}
\feature{economist}\\
\paragraph{\textbf{\texttt{Religion:}}}
\feature{anglican priest}, \feature{bible translator}, \feature{bishop}, \feature{catholic priest}, \feature{christian monk}, \feature{lay theologian}, \feature{monk}, \feature{pastor}, \feature{pope}, \feature{preacher}, \feature{priest}, \feature{theologian}\\
\paragraph{\textbf{\texttt{Military:}}}
\feature{air force officer}, \feature{aircraft pilot}, \feature{commanding officer}, \feature{fighter pilot}, \feature{general officer}, \feature{helicopter pilot}, \feature{intelligence officer}, \feature{naval officer}, \feature{officer of the french navy}, \feature{police officer}, \feature{soldier}, \feature{starship pilot}, \feature{test pilot}\\
\paragraph{\textbf{\texttt{Translation:}}}
\feature{translator}\\
\paragraph{\textbf{\texttt{Philosophy:}}}
\feature{analytic philosopher}, \feature{philosopher}, \feature{philosopher of language}, \feature{philosopher of science}\\
\paragraph{\textbf{\texttt{Polymath:}}}
\feature{polymath}\\
\clearpage
\begin{table*}[t]
    \small\small
    \centering
    \setlength{\tabcolsep}{28pt} 
    \resizebox{\linewidth}{!}{%
        \begin{tabular}{llrcccc}
        \textbf{Dataset} & \textbf{Features}  & \textbf{Coef} & \textbf{SE} & \textbf{Wald (|W|)} & \multicolumn{2}{l}{\textbf{$\mathbb{P}_{Z\sim\mathcal{N}}(|Z| > |W|)$}} \\ 
        \toprule
        
\multirow{21}{*}{\parbox{3.5 cm}{\textbf{\large\nq{} \\  \\  \\ Model~Fit: \\ \\ 78.09\%}}}
	 & \blutext{\lrfeature{\textbf{bias}}}
	 & \blutext{\textbf{+1.964}}
	 & \blutext{\textbf{0.727}}
	 & \blutext{\textbf{2.703}}
	 & \blutext{\textbf{0.0069}}
	 & \blutext{\textbf{\mystar\mystar\mystar}}
	
\\

	 & \negtext{\lrfeature{\textbf{multi\_answers}}}
	 & \negtext{\textbf{-1.893}}
	 & \negtext{\textbf{0.438}}
	 & \negtext{\textbf{4.327}}
	 & \negtext{\textbf{0.0000}}
	 & \negtext{\textbf{\mystar\mystar\mystar\mystar}}
	
\\

	 & \negtext{\lrfeature{\textbf{o\_not\_found}}}
	 & \negtext{\textbf{-1.112}}
	 & \negtext{\textbf{0.514}}
	 & \negtext{\textbf{2.163}}
	 & \negtext{\textbf{0.0305}}
	 & \negtext{\textbf{\mystar\mystar}}
	
\\

	 & \postext{\lrfeature{\textbf{t\_who}}}
	 & \postext{\textbf{+0.773}}
	 & \postext{\textbf{0.280}}
	 & \postext{\textbf{2.764}}
	 & \postext{\textbf{0.0057}}
	 & \postext{\textbf{\mystar\mystar\mystar}}
	
\\

	 & \postext{\lrfeature{o\_law/crime}}
	 & \postext{+0.729}
	 & \postext{0.738}
	 & \postext{0.989}
	 & \postext{0.3227}
	 & \postext{\textbf{}}
	
\\

	 & \negtext{\lrfeature{\textbf{o\_science/tech}}}
	 & \negtext{\textbf{-0.715}}
	 & \negtext{\textbf{0.390}}
	 & \negtext{\textbf{1.832}}
	 & \negtext{\textbf{0.0670}}
	 & \negtext{\textbf{\mystar}}
	
\\

	 & \negtext{\lrfeature{t\_where}}
	 & \negtext{-0.695}
	 & \negtext{0.645}
	 & \negtext{1.078}
	 & \negtext{0.2811}
	 & \negtext{\textbf{}}
	
\\

	 & \negtext{\lrfeature{\textbf{multi\_entities}}}
	 & \negtext{\textbf{-0.678}}
	 & \negtext{\textbf{0.342}}
	 & \negtext{\textbf{1.979}}
	 & \negtext{\textbf{0.0478}}
	 & \negtext{\textbf{\mystar\mystar}}
	
\\

	 & \postext{\lrfeature{t\_what}}
	 & \postext{+0.505}
	 & \postext{0.415}
	 & \postext{1.216}
	 & \postext{0.2240}
	 & \postext{\textbf{}}
	
\\

	 & \negtext{\lrfeature{o\_stage}}
	 & \negtext{-0.444}
	 & \negtext{0.607}
	 & \negtext{0.732}
	 & \negtext{0.4640}
	 & \negtext{\textbf{}}
	
\\

	 & \postext{\lrfeature{g\_female}}
	 & \postext{+0.427}
	 & \postext{0.489}
	 & \postext{0.873}
	 & \postext{0.3829}
	 & \postext{\textbf{}}
	
\\

	 & \postext{\lrfeature{o\_executive}}
	 & \postext{+0.425}
	 & \postext{0.684}
	 & \postext{0.622}
	 & \postext{0.5341}
	 & \postext{\textbf{}}
	
\\

	 & \postext{\lrfeature{\textbf{q\_sim}}}
	 & \postext{\textbf{+0.406}}
	 & \postext{\textbf{0.210}}
	 & \postext{\textbf{1.934}}
	 & \postext{\textbf{0.0531}}
	 & \postext{\textbf{\mystar}}
	
\\

	 & \postext{\lrfeature{\textbf{e\_train\_count}}}
	 & \postext{\textbf{+0.353}}
	 & \postext{\textbf{0.178}}
	 & \postext{\textbf{1.979}}
	 & \postext{\textbf{0.0479}}
	 & \postext{\textbf{\mystar\mystar}}
	
\\

	 & \postext{\lrfeature{o\_others}}
	 & \postext{+0.346}
	 & \postext{0.443}
	 & \postext{0.781}
	 & \postext{0.4346}
	 & \postext{\textbf{}}
	
\\

	 & \postext{\lrfeature{o\_music}}
	 & \postext{+0.310}
	 & \postext{0.259}
	 & \postext{1.200}
	 & \postext{0.2302}
	 & \postext{\textbf{}}
	
\\

	 & \negtext{\lrfeature{o\_fictional}}
	 & \negtext{-0.256}
	 & \negtext{0.445}
	 & \negtext{0.576}
	 & \negtext{0.5644}
	 & \negtext{\textbf{}}
	
\\

	 & \negtext{\lrfeature{o\_politics}}
	 & \negtext{-0.197}
	 & \negtext{0.304}
	 & \negtext{0.646}
	 & \negtext{0.5185}
	 & \negtext{\textbf{}}
	
\\

	 & \postext{\lrfeature{o\_film/tv}}
	 & \postext{+0.142}
	 & \postext{0.226}
	 & \postext{0.629}
	 & \postext{0.5294}
	 & \postext{\textbf{}}
	
\\

	 & \negtext{\lrfeature{t\_when}}
	 & \negtext{-0.134}
	 & \negtext{0.378}
	 & \negtext{0.354}
	 & \negtext{0.7233}
	 & \negtext{\textbf{}}
	
\\

	 & \negtext{\lrfeature{g\_male}}
	 & \negtext{-0.126}
	 & \negtext{0.508}
	 & \negtext{0.249}
	 & \negtext{0.8033}
	 & \negtext{\textbf{}}
	
\\

\midrule
\multirow{18}{*}{\parbox{2.0 cm}{\textbf{\large \qb{} \\  \\  \\ Model~Fit: \\ \\ 71.90\%}}}
	 & \postext{\lrfeature{\textbf{e\_train\_count}}}
	 & \postext{\textbf{+1.922}}
	 & \postext{\textbf{0.269}}
	 & \postext{\textbf{7.144}}
	 & \postext{\textbf{0.0000}}
	 & \postext{\textbf{\mystar\mystar\mystar\mystar}}
	
\\

	 & \blutext{\lrfeature{\textbf{bias}}}
	 & \blutext{\textbf{-1.024}}
	 & \blutext{\textbf{0.291}}
	 & \blutext{\textbf{3.516}}
	 & \blutext{\textbf{0.0004}}
	 & \blutext{\textbf{\mystar\mystar\mystar\mystar}}
	
\\

	 & \negtext{\lrfeature{\textbf{o\_film/tv}}}
	 & \negtext{\textbf{-0.910}}
	 & \negtext{\textbf{0.470}}
	 & \negtext{\textbf{1.934}}
	 & \negtext{\textbf{0.0531}}
	 & \negtext{\textbf{\mystar}}
	
\\

	 & \negtext{\lrfeature{\textbf{multi\_entities}}}
	 & \negtext{\textbf{-0.870}}
	 & \negtext{\textbf{0.165}}
	 & \negtext{\textbf{5.287}}
	 & \negtext{\textbf{0.0000}}
	 & \negtext{\textbf{\mystar\mystar\mystar\mystar}}
	
\\

	 & \negtext{\lrfeature{t\_what}}
	 & \negtext{-0.734}
	 & \negtext{0.457}
	 & \negtext{1.605}
	 & \negtext{0.1085}
	 & \negtext{\textbf{}}
	
\\

	 & \postext{\lrfeature{o\_translation}}
	 & \postext{+0.671}
	 & \postext{1.439}
	 & \postext{0.466}
	 & \postext{0.6412}
	 & \postext{\textbf{}}
	
\\

	 & \negtext{\lrfeature{\textbf{o\_science/tech}}}
	 & \negtext{\textbf{-0.667}}
	 & \negtext{\textbf{0.265}}
	 & \negtext{\textbf{2.522}}
	 & \negtext{\textbf{0.0117}}
	 & \negtext{\textbf{\mystar\mystar}}
	
\\

	 & \negtext{\lrfeature{\textbf{o\_religion}}}
	 & \negtext{\textbf{-0.655}}
	 & \negtext{\textbf{0.362}}
	 & \negtext{\textbf{1.812}}
	 & \negtext{\textbf{0.0700}}
	 & \negtext{\textbf{\mystar}}
	
\\

	 & \postext{\lrfeature{t\_where}}
	 & \postext{+0.545}
	 & \postext{0.615}
	 & \postext{0.885}
	 & \postext{0.3760}
	 & \postext{\textbf{}}
	
\\

	 & \negtext{\lrfeature{o\_history}}
	 & \negtext{-0.484}
	 & \negtext{0.436}
	 & \negtext{1.109}
	 & \negtext{0.2673}
	 & \negtext{\textbf{}}
	
\\

	 & \postext{\lrfeature{\textbf{o\_writing}}}
	 & \postext{\textbf{+0.402}}
	 & \postext{\textbf{0.189}}
	 & \postext{\textbf{2.128}}
	 & \postext{\textbf{0.0334}}
	 & \postext{\textbf{\mystar\mystar}}
	
\\

	 & \negtext{\lrfeature{o\_not\_found}}
	 & \negtext{-0.382}
	 & \negtext{0.349}
	 & \negtext{1.093}
	 & \negtext{0.2744}
	 & \negtext{\textbf{}}
	
\\

	 & \negtext{\lrfeature{o\_fictional}}
	 & \negtext{-0.368}
	 & \negtext{0.540}
	 & \negtext{0.681}
	 & \negtext{0.4957}
	 & \negtext{\textbf{}}
	
\\

	 & \postext{\lrfeature{\textbf{t\_who}}}
	 & \postext{\textbf{+0.363}}
	 & \postext{\textbf{0.183}}
	 & \postext{\textbf{1.990}}
	 & \postext{\textbf{0.0466}}
	 & \postext{\textbf{\mystar\mystar}}
	
\\

	 & \postext{\lrfeature{o\_artist}}
	 & \postext{+0.194}
	 & \postext{0.268}
	 & \postext{0.721}
	 & \postext{0.4709}
	 & \postext{\textbf{}}
	
\\

	 & \postext{\lrfeature{o\_music}}
	 & \postext{+0.123}
	 & \postext{0.249}
	 & \postext{0.492}
	 & \postext{0.6225}
	 & \postext{\textbf{}}
	
\\

	 & \negtext{\lrfeature{o\_politics}}
	 & \negtext{-0.113}
	 & \negtext{0.202}
	 & \negtext{0.560}
	 & \negtext{0.5756}
	 & \negtext{\textbf{}}
	
\\

	 & \negtext{\lrfeature{o\_philosophy}}
	 & \negtext{-0.082}
	 & \negtext{0.259}
	 & \negtext{0.315}
	 & \negtext{0.7530}
	 & \negtext{\textbf{}}
	
\\

\midrule
\multirow{22}{*}{\parbox{2.0 cm}{\textbf{\large\triviaqa{} \\  \\  \\ Model~Fit: \\ \\  60.18\%}}}
	 & \blutext{\lrfeature{\textbf{bias}}}
	 & \blutext{\textbf{-1.066}}
	 & \blutext{\textbf{0.114}}
	 & \blutext{\textbf{9.353}}
	 & \blutext{\textbf{0.0000}}
	 & \blutext{\textbf{\mystar\mystar\mystar\mystar}}
	
\\

	 & \negtext{\lrfeature{o\_education}}
	 & \negtext{-0.665}
	 & \negtext{0.534}
	 & \negtext{1.245}
	 & \negtext{0.2132}
	 & \negtext{\textbf{}}
	
\\

	 & \postext{\lrfeature{o\_economics}}
	 & \postext{+0.503}
	 & \postext{0.579}
	 & \postext{0.869}
	 & \postext{0.3848}
	 & \postext{\textbf{}}
	
\\

	 & \negtext{\lrfeature{o\_philosophy}}
	 & \negtext{-0.469}
	 & \negtext{0.342}
	 & \negtext{1.370}
	 & \negtext{0.1706}
	 & \negtext{\textbf{}}
	
\\

	 & \negtext{\lrfeature{\textbf{o\_religion}}}
	 & \negtext{\textbf{-0.443}}
	 & \negtext{\textbf{0.255}}
	 & \negtext{\textbf{1.738}}
	 & \negtext{\textbf{0.0822}}
	 & \negtext{\textbf{\mystar}}
	
\\

	 & \postext{\lrfeature{\textbf{o\_law/crime}}}
	 & \postext{\textbf{+0.412}}
	 & \postext{\textbf{0.218}}
	 & \postext{\textbf{1.890}}
	 & \postext{\textbf{0.0588}}
	 & \postext{\textbf{\mystar}}
	
\\

	 & \postext{\lrfeature{\textbf{multi\_answers}}}
	 & \postext{\textbf{+0.341}}
	 & \postext{\textbf{0.024}}
	 & \postext{\textbf{14.090}}
	 & \postext{\textbf{0.0000}}
	 & \postext{\textbf{\mystar\mystar\mystar\mystar}}
	
\\

	 & \postext{\lrfeature{\textbf{t\_who}}}
	 & \postext{\textbf{+0.230}}
	 & \postext{\textbf{0.129}}
	 & \postext{\textbf{1.778}}
	 & \postext{\textbf{0.0754}}
	 & \postext{\textbf{\mystar}}
	
\\

	 & \postext{\lrfeature{o\_science/tech}}
	 & \postext{+0.219}
	 & \postext{0.151}
	 & \postext{1.452}
	 & \postext{0.1466}
	 & \postext{\textbf{}}
	
\\

	 & \negtext{\lrfeature{o\_stage}}
	 & \negtext{-0.212}
	 & \negtext{0.257}
	 & \negtext{0.824}
	 & \negtext{0.4098}
	 & \negtext{\textbf{}}
	
\\

	 & \negtext{\lrfeature{\textbf{o\_politics}}}
	 & \negtext{\textbf{-0.208}}
	 & \negtext{\textbf{0.095}}
	 & \negtext{\textbf{2.177}}
	 & \negtext{\textbf{0.0295}}
	 & \negtext{\textbf{\mystar\mystar}}
	
\\

	 & \postext{\lrfeature{\textbf{o\_writing}}}
	 & \postext{\textbf{+0.192}}
	 & \postext{\textbf{0.098}}
	 & \postext{\textbf{1.955}}
	 & \postext{\textbf{0.0506}}
	 & \postext{\textbf{\mystar}}
	
\\

	 & \negtext{\lrfeature{o\_sports}}
	 & \negtext{-0.183}
	 & \negtext{0.117}
	 & \negtext{1.570}
	 & \negtext{0.1164}
	 & \negtext{\textbf{}}
	
\\

	 & \postext{\lrfeature{o\_film/tv}}
	 & \postext{+0.150}
	 & \postext{0.092}
	 & \postext{1.626}
	 & \postext{0.1039}
	 & \postext{\textbf{}}
	
\\

	 & \negtext{\lrfeature{t\_when}}
	 & \negtext{-0.146}
	 & \negtext{0.299}
	 & \negtext{0.488}
	 & \negtext{0.6258}
	 & \negtext{\textbf{}}
	
\\

	 & \negtext{\lrfeature{o\_not\_found}}
	 & \negtext{-0.143}
	 & \negtext{0.205}
	 & \negtext{0.699}
	 & \negtext{0.4845}
	 & \negtext{\textbf{}}
	
\\

	 & \negtext{\lrfeature{o\_executive}}
	 & \negtext{-0.130}
	 & \negtext{0.207}
	 & \negtext{0.625}
	 & \negtext{0.5323}
	 & \negtext{\textbf{}}
	
\\

	 & \negtext{\lrfeature{o\_others}}
	 & \negtext{-0.101}
	 & \negtext{0.152}
	 & \negtext{0.665}
	 & \negtext{0.5058}
	 & \negtext{\textbf{}}
	
\\

	 & \negtext{\lrfeature{multi\_entities}}
	 & \negtext{-0.099}
	 & \negtext{0.077}
	 & \negtext{1.290}
	 & \negtext{0.1971}
	 & \negtext{\textbf{}}
	
\\

	 & \postext{\lrfeature{e\_train\_count}}
	 & \postext{+0.081}
	 & \postext{0.075}
	 & \postext{1.079}
	 & \postext{0.2804}
	 & \postext{\textbf{}}
	
\\

	 & \postext{\lrfeature{o\_music}}
	 & \postext{+0.022}
	 & \postext{0.097}
	 & \postext{0.223}
	 & \postext{0.8237}
	 & \postext{\textbf{}}
	
\\

	 & \postext{\lrfeature{t\_what}}
	 & \postext{+0.018}
	 & \postext{0.134}
	 & \postext{0.132}
	 & \postext{0.8947}
	 & \postext{\textbf{}}
	
\\

\bottomrule
        \end{tabular}
    }
    
\caption{\label{tab:logistic-regression-appendix}Influential features revealed through Logistic Regression Analysis (Sec~\ref{subsec:logistic-regression}) over the demographic \democol{}s deemed significant through the $\chi^2$ test (Figure~\ref{fig:accuracies}). We report the highly influential features with significance of $p$-value < $0.1$, both positive (blue) and negative (red), and \textbf{bold} the highly significant ones ($p$-value < $0.05$). Number of \mystar{} in the last column represents the significance level of that feature.}
\end{table*}

\section{Logistic Regression features.}
\label{appendix:logistic-regression}
This section enlists a full set of features used for the logistic regression analysis after feature reduction, each with their coefficients, standard error, Wald Statistic and significance level in Table~\ref{tab:logistic-regression-appendix}. We also describe the templates and the implementation details of the features using in our logistic regression analysis (Section~\ref{subsec:logistic-regression}) in Appendix~\ref{appendix:feature-implementation}, and finally enlist some randomly sampled examples both from \nq{} and \triviaqa{} datasets in Appendix~\ref{appendix:multi-answers-examples} to show how \lrfeature{multi\_answers} feature has disparate effects on them.

\subsection{Implementation of Logistic Regression features}
\label{appendix:feature-implementation}
\begin{itemize}
  \item \lrfeature{q\_sim}: For closed-domain \qa{} tasks like \nq{} and \squad{}, this feature measures (sim)ilarity between (q)uestion text and evidence sentence---the sentence from the evidence passage which contains the answer text---using Jaccard similarity over unigram tokens~\cite{sugawara-18}. Since we do not include \squad{} in our logistic regression analysis (Section~\ref{subsec:logistic-regression}, this feature is only relevant for \nq{}.
  
  \item \lrfeature{e\_train\_count}: This binary feature represents if distinct (e)ntities appearing in a \qa{} example (through the approach described in Section~\ref{sec:mapping}) appears more than twice in the particular dataset's training fold. We avoid logarithm here as even the log frequency for some commonly occurring entities exceeds the expected feature value range.
  
  \item \lrfeature{t\_wh*}: This represents the features that captures the expected entity type of the answer: \lrfeature{t\_who}, \lrfeature{t\_what}, \lrfeature{t\_where}, \lrfeature{t\_when}. Each binary feature captures if the particular \lrfeature{"wh*"} word appears in the first ten (t)okens of the question text.\footnote{\qb{} questions often start with ``For 10 points, name this writer \textit{who}...''}
  
  \item \lrfeature{multi\_entities}: For number of linked person-entities in a example as described in Section~\ref{sec:mapping} as $n$, this feature is $log_2(n)$. Hence, this feature is 0 for example with just single person entity.
  
  \item \lrfeature{multi\_answers}: For number of gold-answers annotated in a example as $n$, this feature is $log_2(n)$. Hence, this feature is 0 for example with just answer.
  
  \item \lrfeature{g\_*}: Binary demographic feature signaling the presence of the (g)ender characterized by the feature. For instance, \lrfeature{g\_female} signals if the question is about a female person.
  
  \item \lrfeature{o\_*}: Binary demographic feature signaling the presence of the occupation (or profession) as characterized by the feature. For instance, \lrfeature{o\_writer} signals if the question is about a writer.
\end{itemize}
\subsection{Examples with \lrfeature{multi\_answers} feature}
\label{appendix:multi-answers-examples}
In the Logistic Regression analysis (Section~\ref{subsec:logistic-regression}), we create two features: \lrfeature{multi\_answers} and \lrfeature{multi\_entities}. Former captures the presence of multiple gold answers to the question in a given example, while latter signals presence of multiple person entities — all in either the answers, the question text or the document title for a given example. While \lrfeature{multi\_entities} has consistent negative co-relation with model correctness (Appendix \ref{appendix:logistic-regression}), \lrfeature{multi\_answers} has a disparate effect.
Though it signals towards incorrectly answered examples in \nq{}, it has a statistically significant positive correlation with model correctness for \triviaqa{} examples. Going through the examples, it reveals that \triviaqa{} uses multiple answers to give alternate formulations of an answer, which aids machine reading, while multiple \nq{} answers are often a sign of question ambiguity~\cite{min-etal-2020-ambigqa}.

To demonstrate that, we enlist here examples from development fold of both \nq{} (Appendix~\ref{appendix:nq-examples}) and \triviaqa{} (Appendix~\ref{appendix:trivia-qa-examples}) that have multiple gold answers.

\subsubsection{\nq{} examples with multiple answers:}
\label{appendix:nq-examples}
\setlength{\parindent}{0em}
\tiny{\setlength{\parindent}{0cm}
\textbf{id: \texttt{-4135209844918483842}} \\
Q: \textit{who carried the us flag in the 2014 olympics} \\
A: \entity{Todd Lodwick} \\
A: \entity{Julie Chu}}

\tiny{\setlength{\parindent}{0cm}
\textbf{id: \texttt{8838716539218945006}} \\
Q: \textit{who says that which we call a rose} \\
A: \entity{William Shakespeare} \\
A: \entity{Juliet}}

\tiny{\setlength{\parindent}{0cm}
\textbf{id: \texttt{-6197052503812142206}} \\
Q: \textit{who has won the most superbowls as a player} \\
A: \entity{Charles Haley} \\
A: \entity{Tom Brady}}

\tiny{\setlength{\parindent}{0cm}
\textbf{id: \texttt{-2840415450119119129}} \\
Q: \textit{who started the guinness book of world records} \\
A: \entity{Hugh Beaver} \\
A: \entity{Norris and Ross McWhirter}}

\tiny{\setlength{\parindent}{0cm}
\textbf{id: \texttt{6997422338613101186}} \\
Q: \textit{who played the nurse on andy griffith show} \\
A: \entity{Langdon} \\
A: \entity{Julie Adams}}

\tiny{\setlength{\parindent}{0cm}
\textbf{id: \texttt{-7064677612340044331}} \\
Q: \textit{who wrote the song if i were a boy} \\
A: \entity{BC Jean} \\
A: \entity{Toby Gad}}

\tiny{\setlength{\parindent}{0cm}
\textbf{id: \texttt{3248410603422198181}} \\
Q: \textit{who conducted the opening concert at carnegie hall} \\
A: \entity{Walter Damrosch} \\
A: \entity{Pyotr Ilyich Tchaikovsky}}

\tiny{\setlength{\parindent}{0cm}
\textbf{id: \texttt{-3772952199709196386}} \\
Q: \textit{who founded amazon where is the headquarters of amazon} \\
A: \entity{founded by Jeff Bezos} \\
A: \entity{based in Seattle , Washington}}

\tiny{\setlength{\parindent}{0cm}
\textbf{id: \texttt{4053461415821443645}} \\
Q: \textit{who wrote song what a friend we have in jesus} \\
A: \entity{Joseph M. Scriven} \\
A: \entity{Charles Crozat Converse}}

\tiny{\setlength{\parindent}{0cm}
\textbf{id: \texttt{-5670674709553776773}} \\
Q: \textit{who sings the theme song for the proud family} \\
A: \entity{Solange Knowles} \\
A: \entity{Solange Knowles}}

\tiny{\setlength{\parindent}{0cm}
\textbf{id: \texttt{2978779480736570480}} \\
Q: \textit{days of our lives cast doug and julie} \\
A: \entity{Bill Hayes} \\
A: \entity{Susan Seaforth}}

\tiny{\setlength{\parindent}{0cm}
\textbf{id: \texttt{6173192803639008655}} \\
Q: \textit{who has appeared in the most royal rumbles} \\
A: \entity{Isaac Yankem / `` Diesel '' / Kane} \\
A: \entity{Shawn Michaels}}

\tiny{\setlength{\parindent}{0cm}
\textbf{id: \texttt{7561389892504775773}} \\
Q: \textit{who wrote the song stop the world and let me off} \\
A: \entity{Carl Belew} \\
A: \entity{W.S. Stevenson}}

\tiny{\setlength{\parindent}{0cm}
\textbf{id: \texttt{-8366545547296627039}} \\
Q: \textit{who wrote the song photograph by ringo starr} \\
A: \entity{Ringo Starr} \\
A: \entity{George Harrison}}

\tiny{\setlength{\parindent}{0cm}
\textbf{id: \texttt{-5674327280636928690}} \\
Q: \textit{who sings you're welcome in moana credits} \\
A: \entity{Lin - Manuel Miranda} \\
A: \entity{Jordan Fisher}}

\tiny{\setlength{\parindent}{0cm}
\textbf{id: \texttt{-2432292250757146771}} \\
Q: \textit{who wrote the song i hate you i love you} \\
A: \entity{Garrett Nash} \\
A: \entity{Olivia O'Brien}}

\tiny{\setlength{\parindent}{0cm}
\textbf{id: \texttt{-3632974700795137148}} \\
Q: \textit{who is the owner of reading football club} \\
A: \entity{Xiu Li Dai} \\
A: \entity{Yongge Dai}}

\tiny{\setlength{\parindent}{0cm}
\textbf{id: \texttt{7163132803738849961}} \\
Q: \textit{who played guitar on my guitar gently weeps} \\
A: \entity{Eric Clapton} \\
A: \entity{George Harrison}}

\tiny{\setlength{\parindent}{0cm}
\textbf{id: \texttt{1318031841813121387}} \\
Q: \textit{who sang the theme song to that 70s show} \\
A: \entity{Todd Griffin} \\
A: \entity{Cheap Trick}}

\tiny{\setlength{\parindent}{0cm}
\textbf{id: \texttt{1393634180793653648}} \\
Q: \textit{who came up with the initial concept of protons and neutrons} \\
A: \entity{Werner Heisenberg} \\
A: \entity{Dmitri Ivanenko}}

\tiny{\setlength{\parindent}{0cm}
\textbf{id: \texttt{9134704289334516617}} \\
Q: \textit{who missed the plane the day the music died} \\
A: \entity{Waylon Jennings} \\
A: \entity{Tommy Allsup}}

\tiny{\setlength{\parindent}{0cm}
\textbf{id: \texttt{8466196474705624263}} \\
Q: \textit{who was running as vice president in 1984} \\
A: \entity{Congresswoman Ferraro} \\
A: \entity{Vice President George H.W. Bush}}

\tiny{\setlength{\parindent}{0cm}
\textbf{id: \texttt{5579013873387598720}} \\
Q: \textit{who has won the canada open women's doubles} \\
A: \entity{Mayu Matsumoto} \\
A: \entity{Wakana Nagahara}}

\tiny{\setlength{\parindent}{0cm}
\textbf{id: \texttt{5584540254904933863}} \\
Q: \textit{who sang what are we doing in love} \\
A: \entity{Dottie West} \\
A: \entity{Kenny Rogers}}

\tiny{\setlength{\parindent}{0cm}
\textbf{id: \texttt{-8677459248394445003}} \\
Q: \textit{who is hosting e live from the red carpet} \\
A: \entity{Ryan Seacrest} \\
A: \entity{Giuliana Rancic}}

\tiny{\setlength{\parindent}{0cm}
\textbf{id: \texttt{-1342189058950802702}} \\
Q: \textit{who made the poppies at tower of london} \\
A: \entity{Paul Cummins} \\
A: \entity{Tom Piper}}

\tiny{\setlength{\parindent}{0cm}
\textbf{id: \texttt{6014950976264156000}} \\
Q: \textit{who sang never gonna let you go} \\
A: \entity{Joe Pizzulo} \\
A: \entity{Leeza Miller}}

\tiny{\setlength{\parindent}{0cm}
\textbf{id: \texttt{-8052136860650205450}} \\
Q: \textit{who wrote the song rainy days and mondays} \\
A: \entity{Roger Nichols} \\
A: \entity{Paul Williams}}

\tiny{\setlength{\parindent}{0cm}
\textbf{id: \texttt{7903911150166287814}} \\
Q: \textit{what position did doug peterson play in the nfl} \\
A: \entity{quarterback} \\
A: \entity{holder on placekicks}}

\tiny{\setlength{\parindent}{0cm}
\textbf{id: \texttt{583026970021621830}} \\
Q: \textit{who invented the first home video security system} \\
A: \entity{Marie Van Brittan Brown} \\
A: \entity{her husband Albert Brown}}

\tiny{\setlength{\parindent}{0cm}
\textbf{id: \texttt{5427679691711111925}} \\
Q: \textit{who were the two mathematicians that invented calculus} \\
A: \entity{Isaac Newton} \\
A: \entity{Gottfried Leibniz}}

\tiny{\setlength{\parindent}{0cm}
\textbf{id: \texttt{-9163844183450408581}} \\
Q: \textit{nba record for most double doubles in a season} \\
A: \entity{Tim Duncan leads the National Basketball Association ( NBA ) in the points - rebounds combination with 840} \\
A: \entity{John Stockton leads the points - assists combination with 714}}

\tiny{\setlength{\parindent}{0cm}
\textbf{id: \texttt{-8109367537690343895}} \\
Q: \textit{who were the twins that played for kentucky} \\
A: \entity{Andrew Michael Harrison} \\
A: \entity{Aaron Harrison}}

\tiny{\setlength{\parindent}{0cm}
\textbf{id: \texttt{4784420206031467202}} \\
Q: \textit{who wrote he ain't heavy he's my brother lyrics} \\
A: \entity{Bobby Scott} \\
A: \entity{Bob Russell}}

\tiny{\setlength{\parindent}{0cm}
\textbf{id: \texttt{4136958282795887427}} \\
Q: \textit{who opens the church of the holy sepulchre} \\
A: \entity{the Nusaybah family} \\
A: \entity{the Joudeh Al - Goudia family}}

\tiny{\setlength{\parindent}{0cm}
\textbf{id: \texttt{-2610209560699528896}} \\
Q: \textit{who is the writer of 50 shades of grey} \\
A: \entity{Erika Mitchell Leonard} \\
A: \entity{E.L. James}}

\tiny{\setlength{\parindent}{0cm}
\textbf{id: \texttt{8968036245733884389}} \\
Q: \textit{when did stephen curry won the mvp award} \\
A: \entity{2015 ,} \\
A: \entity{2015}}

\tiny{\setlength{\parindent}{0cm}
\textbf{id: \texttt{-1899514742808499173}} \\
Q: \textit{who are nominated for president of india 2017} \\
A: \entity{Ram Nath Kovind} \\
A: \entity{Meira Kumar}}

\tiny{\setlength{\parindent}{0cm}
\textbf{id: \texttt{-3019484115332998709}} \\
Q: \textit{what movie is count on me by bruno mars in} \\
A: \entity{A Turtle 's Tale : Sammy 's Adventures} \\
A: \entity{Diary of a Wimpy Kid : The Long Haul}}

\tiny{\setlength{\parindent}{0cm}
\textbf{id: \texttt{810060125994185205}} \\
Q: \textit{who was the first to say i'm going to disney world} \\
A: \entity{Dick Rutan} \\
A: \entity{Jeana Yeager}}

\tiny{\setlength{\parindent}{0cm}
\textbf{id: \texttt{339027965927992295}} \\
Q: \textit{who sings the whiskey ain't workin anymore} \\
A: \entity{Travis Tritt} \\
A: \entity{Marty Stuart}}

\tiny{\setlength{\parindent}{0cm}
\textbf{id: \texttt{5995814638252489040}} \\
Q: \textit{who played scotty baldwins father on general hospital} \\
A: \entity{Peter Hansen} \\
A: \entity{Ross Elliott}}

\tiny{\setlength{\parindent}{0cm}
\textbf{id: \texttt{3723628014502752965}} \\
Q: \textit{who wrote cant get you out of my head lyrics} \\
A: \entity{Cathy Dennis} \\
A: \entity{Rob Davis}}

\tiny{\setlength{\parindent}{0cm}
\textbf{id: \texttt{3886074985605209321}} \\
Q: \textit{who sings find out who your friends are with tracy lawrence} \\
A: \entity{Tim McGraw} \\
A: \entity{Kenny Chesney}}

\tiny{\setlength{\parindent}{0cm}
\textbf{id: \texttt{3624266518328727040}} \\
Q: \textit{who invented the printing press and what year} \\
A: \entity{Johannes Gutenberg} \\
A: \entity{circa 1439}}

\tiny{\setlength{\parindent}{0cm}
\textbf{id: \texttt{-4951004239400083779}} \\
Q: \textit{who plays chris grandy in 13 going on 30} \\
A: \entity{Jim Gaffigan} \\
A: \entity{Alex Black}}

\tiny{\setlength{\parindent}{0cm}
\textbf{id: \texttt{2672721743911117185}} \\
Q: \textit{who developed a set of postulates to prove that specific microorganisms cause disease} \\
A: \entity{Robert Koch} \\
A: \entity{Friedrich Loeffler}}

\tiny{\setlength{\parindent}{0cm}
\textbf{id: \texttt{2166092801797515500}} \\
Q: \textit{who is the director of taarak mehta ka ooltah chashmah} \\
A: \entity{Harshad Joshi} \\
A: \entity{Malav Suresh Rajda}}

\tiny{\setlength{\parindent}{0cm}
\textbf{id: \texttt{-3389723371168293793}} \\
Q: \textit{who has the most olympic medals in figure skating} \\
A: \entity{Tessa Virtue} \\
A: \entity{Scott Moir}}

\tiny{\setlength{\parindent}{0cm}
\textbf{id: \texttt{-8391680223788694572}} \\
Q: \textit{who wrote if i were a boy reba or beyonce} \\
A: \entity{BC Jean} \\
A: \entity{Toby Gad}}

\tiny{\setlength{\parindent}{0cm}
\textbf{id: \texttt{1070572237499172286}} \\
Q: \textit{who wrote the song after you've gone} \\
A: \entity{Turner Layton} \\
A: \entity{Henry Creamer}}

\tiny{\setlength{\parindent}{0cm}
\textbf{id: \texttt{2343902375984110832}} \\
Q: \textit{who does the voice of mickey mouse on mickey mouse clubhouse} \\
A: \entity{Wayne Allwine} \\
A: \entity{Bret Iwan}}

\tiny{\setlength{\parindent}{0cm}
\textbf{id: \texttt{7013863939803495694}} \\
Q: \textit{who sings love me tender in princess diaries 2} \\
A: \entity{Norah Jones} \\
A: \entity{Adam Levy}}

\tiny{\setlength{\parindent}{0cm}
\textbf{id: \texttt{4925057086725798331}} \\
Q: \textit{who wrote yakkity yak don't talk back} \\
A: \entity{Jerry Leiber} \\
A: \entity{Mike Stoller}}

\tiny{\setlength{\parindent}{0cm}
\textbf{id: \texttt{647605647914971565}} \\
Q: \textit{who wrote lyrics for phantom of the opera} \\
A: \entity{Charles Hart} \\
A: \entity{Richard Stilgoe}}

\tiny{\setlength{\parindent}{0cm}
\textbf{id: \texttt{-6371603500131574271}} \\
Q: \textit{who sings somebody's watching me with michael jackson} \\
A: \entity{Rockwell} \\
A: \entity{Jermaine Jackson}}

\tiny{\setlength{\parindent}{0cm}
\textbf{id: \texttt{-4036503601399675973}} \\
Q: \textit{when did michael jordan return to the nba} \\
A: \entity{1995} \\
A: \entity{2001}}

\tiny{\setlength{\parindent}{0cm}
\textbf{id: \texttt{4323871331649279373}} \\
Q: \textit{who invented the printing press and in what year} \\
A: \entity{Johannes Gutenberg} \\
A: \entity{1440}}

\tiny{\setlength{\parindent}{0cm}
\textbf{id: \texttt{7234277123646852447}} \\
Q: \textit{who sings war don't let me down} \\
A: \entity{American production duo The Chainsmokers} \\
A: \entity{vocals of American singer Daya}}

\tiny{\setlength{\parindent}{0cm}
\textbf{id: \texttt{4245798066923223457}} \\
Q: \textit{who has the most all star mvp awards} \\
A: \entity{Bob Pettit} \\
A: \entity{Kobe Bryant}}

\tiny{\setlength{\parindent}{0cm}
\textbf{id: \texttt{-3585157729928173881}} \\
Q: \textit{who plays hulk in the thor and avengers series of movies} \\
A: \entity{Fred Tatasciore} \\
A: \entity{Rick D. Wasserman}}

\tiny{\setlength{\parindent}{0cm}
\textbf{id: \texttt{-7892904540301629325}} \\
Q: \textit{who wrote the song going to kansas city} \\
A: \entity{Jerry Leiber} \\
A: \entity{Mike Stoller}}

\tiny{\setlength{\parindent}{0cm}
\textbf{id: \texttt{1838851770314085590}} \\
Q: \textit{who plays sheila carter on the bold and the beautiful} \\
A: \entity{Kimberlin Brown} \\
A: \entity{Michelle Stafford}}

\subsubsection{\triviaqa{} multi-answer examples:}
\label{appendix:trivia-qa-examples}
\small{We randomly sample 100 examples from \triviaqa{} where questions had multiple answers.}
\\ 
\setlength{\parindent}{0cm}

\tiny{\setlength{\parindent}{0cm}
\textbf{id: \texttt{sfq\_6110}} \\
Q: \textit{On which island in the North Sea did both St Aidan and St Cuthbert live?} \\
A: \entity{Lindisfarne } \\
A: \entity{LINDISFARNE}}

\tiny{\setlength{\parindent}{0cm}
\textbf{id: \texttt{tc\_1008}} \\
Q: \textit{To the nearest two, how many tennis Grand Slam titles did Jimmy Connors win?} \\
A: \entity{10} \\
A: \entity{ten}}

\tiny{\setlength{\parindent}{0cm}
\textbf{id: \texttt{sfq\_26211}} \\
Q: \textit{In the TV series Doctor Who, who was the creator of the Daleks and arch enemy of the Doctor?} \\
A: \entity{Davros} \\
A: \entity{Creator of the Daleks}}

\tiny{\setlength{\parindent}{0cm}
\textbf{id: \texttt{sfq\_22212}} \\
Q: \textit{In which book of the bible is the story of Samson and Delilah?} \\
A: \entity{Judge (disambiguation)} \\
A: \entity{Judges}}

\tiny{\setlength{\parindent}{0cm}
\textbf{id: \texttt{bt\_1538}} \\
Q: \textit{What is cartoon character Mr. Magoo's first name} \\
A: \entity{Quincy (disambiguation)} \\
A: \entity{Quincy}}

\tiny{\setlength{\parindent}{0cm}
\textbf{id: \texttt{qz\_2444}} \\
Q: \textit{What is Robin Williams character called in Good Morning Vietnam?} \\
A: \entity{Adrian}}

\tiny{\setlength{\parindent}{0cm}
\textbf{id: \texttt{sfq\_22693}} \\
Q: \textit{What was the first name of the jazz trombonist Kid Ory?} \\
A: \entity{Eadweard} \\
A: \entity{Edward}}

\tiny{\setlength{\parindent}{0cm}
\textbf{id: \texttt{qw\_1606}} \\
Q: \textit{Which of Queen Elizabeth's children is the lowest in succession to (i.e. furthest away from) the throne?} \\
A: \entity{Anne} \\
A: \entity{Ann (name)}}

\tiny{\setlength{\parindent}{0cm}
\textbf{id: \texttt{odql\_5503}} \\
Q: \textit{"Which radio comedian's catchphrase was ""daft as a brush""?"} \\
A: \entity{KEN PLATT} \\
A: \entity{Ken Platt}}

\tiny{\setlength{\parindent}{0cm}
\textbf{id: \texttt{qg\_2992}} \\
Q: \textit{According to Sammy Haggar, what can't he drive?} \\
A: \entity{55} \\
A: \entity{fifty-five}}

\tiny{\setlength{\parindent}{0cm}
\textbf{id: \texttt{qf\_3440}} \\
Q: \textit{What was Grace Darling's father's job?} \\
A: \entity{Lighthouse-keeper} \\
A: \entity{Lighthouse keeper}}

\tiny{\setlength{\parindent}{0cm}
\textbf{id: \texttt{qw\_12369}} \\
Q: \textit{"What year did Jean-Francois Champollion publish the first correct translation of Egyptian hieroglyphs from the Rosetta Stone, the Roman Catholic Church take Galileo Galilei's ""Dialogue"" off their list of banned books, and Britain repeal the death penalty for over 100 crimes?"} \\
A: \entity{one thousand, eight hundred and twenty-two} \\
A: \entity{1822}}

\tiny{\setlength{\parindent}{0cm}
\textbf{id: \texttt{qw\_4143}} \\
Q: \textit{What is the title of the most famous painting by Franz Hals?} \\
A: \entity{Laughing Cavalier} \\
A: \entity{The Laughing Cavalier}}

\tiny{\setlength{\parindent}{0cm}
\textbf{id: \texttt{qb\_2647}} \\
Q: \textit{What is the title of the 1944 film starring Barbara Stanwyck as the wife who seduces an insurance salesman into killing her husband?} \\
A: \entity{Double indemnity (disambiguation)} \\
A: \entity{Double Indemnity}}

\tiny{\setlength{\parindent}{0cm}
\textbf{id: \texttt{sfq\_22920}} \\
Q: \textit{Who was the choreographer of the dance troupe Hot Gossip?} \\
A: \entity{Arlene Philips} \\
A: \entity{Arlene Phillips}}

\tiny{\setlength{\parindent}{0cm}
\textbf{id: \texttt{tc\_719}} \\
Q: \textit{River Phoenix died during the making of which movie?} \\
A: \entity{Dark Blood (film)} \\
A: \entity{Dark Blood}}

\tiny{\setlength{\parindent}{0cm}
\textbf{id: \texttt{sfq\_19457}} \\
Q: \textit{Who won the first ever boxing gold for women? She shares her surname with two US Presidents.} \\
A: \entity{Nicola Adams} \\
A: \entity{Adams, Nicola}}

\tiny{\setlength{\parindent}{0cm}
\textbf{id: \texttt{sfq\_8996}} \\
Q: \textit{Actor Norman Painting died in November 2009, which part in a log running radio series did he make his own?} \\
A: \entity{PHIL ARCHER} \\
A: \entity{Phil Archer}}

\tiny{\setlength{\parindent}{0cm}
\textbf{id: \texttt{dpql\_6111}} \\
Q: \textit{Which Jersey-born actor played Superman in Man of Steel?} \\
A: \entity{Henry Cavill} \\
A: \entity{Henry William Dalgliesh Cavill}}

\tiny{\setlength{\parindent}{0cm}
\textbf{id: \texttt{qz\_2135}} \\
Q: \textit{Name the game show, presented by Leslie Grantham and Melinda Messenger, where contestants were set physical and mental challenges?} \\
A: \entity{Fort Boyard (disambiguation)} \\
A: \entity{Fort Boyard}}

\tiny{\setlength{\parindent}{0cm}
\textbf{id: \texttt{odql\_3205}} \\
Q: \textit{Who wrote the novel 'The Beach' on which the film was based?} \\
A: \entity{Alex Garland} \\
A: \entity{ALEX GARLAND}}

\tiny{\setlength{\parindent}{0cm}
\textbf{id: \texttt{qz\_2999}} \\
Q: \textit{In what year did Edward Vlll abdicate?} \\
A: \entity{one thousand, nine hundred and thirty-six} \\
A: \entity{1936}}

\tiny{\setlength{\parindent}{0cm}
\textbf{id: \texttt{tc\_723}} \\
Q: \textit{Which artist David was born in Bradford UK?} \\
A: \entity{Hockney} \\
A: \entity{David Hockney}}

\tiny{\setlength{\parindent}{0cm}
\textbf{id: \texttt{odql\_3708}} \\
Q: \textit{Three Liverpool players were in the 1966 England World Cup winning squad. Roger Hunt and Ian Callaghan were two – who was the third?} \\
A: \entity{Gerry Byrne} \\
A: \entity{Gerry Byrne (disambiguation)}}

\tiny{\setlength{\parindent}{0cm}
\textbf{id: \texttt{qw\_11151}} \\
Q: \textit{Which artist has a daughter and two sons with Jane Asher, whom he married in 1981?} \\
A: \entity{Gerald Anthony Scarfe} \\
A: \entity{Gerald Scarfe}}

\tiny{\setlength{\parindent}{0cm}
\textbf{id: \texttt{qb\_3652}} \\
Q: \textit{Who wrote the novel ‘The Eagle Has landed’?} \\
A: \entity{Harry Patterson} \\
A: \entity{Jack Higgins}}

\tiny{\setlength{\parindent}{0cm}
\textbf{id: \texttt{odql\_14683}} \\
Q: \textit{Who presents the BBC quiz show ‘Perfection’?} \\
A: \entity{Nick Knowles} \\
A: \entity{NICK KNOWLES}}

\tiny{\setlength{\parindent}{0cm}
\textbf{id: \texttt{sfq\_9464}} \\
Q: \textit{Who succeeded Brian Epstein as manager of The Beatles?} \\
A: \entity{Allan Klein} \\
A: \entity{Allen Klein}}

\tiny{\setlength{\parindent}{0cm}
\textbf{id: \texttt{wh\_2615}} \\
Q: \textit{In which year did both T-Rex's Marc Bolan and Elvis Presley die ?} \\
A: \entity{1977} \\
A: \entity{one thousand, nine hundred and seventy-seven}}

\tiny{\setlength{\parindent}{0cm}
\textbf{id: \texttt{sfq\_9018}} \\
Q: \textit{Who played Hotlips Houlihan in the 1972 film MASH?} \\
A: \entity{Sally Kellerman} \\
A: \entity{SALLY KELLERMAN}}

\tiny{\setlength{\parindent}{0cm}
\textbf{id: \texttt{qz\_1516}} \\
Q: \textit{Who bought Chelsea football club for £1 in 1982?} \\
A: \entity{Ken Bates} \\
A: \entity{Kenneth Bates}}

\tiny{\setlength{\parindent}{0cm}
\textbf{id: \texttt{sfq\_23289}} \\
Q: \textit{What was the middle name of the author William Thackeray?} \\
A: \entity{Makepeace} \\
A: \entity{MAKEPEACE}}

\tiny{\setlength{\parindent}{0cm}
\textbf{id: \texttt{sfq\_7589}} \\
Q: \textit{What was the name of the older brother of Henry 8th?} \\
A: \entity{Arthur} \\
A: \entity{Arthur (name)}}

\tiny{\setlength{\parindent}{0cm}
\textbf{id: \texttt{odql\_10316}} \\
Q: \textit{Which actor played 'Hadley', in the TV series of the same name?} \\
A: \entity{GERALD HARPER} \\
A: \entity{Gerald Harper}}

\tiny{\setlength{\parindent}{0cm}
\textbf{id: \texttt{sfq\_12933}} \\
Q: \textit{Operation Barbarossa, Hitler invades Russia.} \\
A: \entity{one thousand, nine hundred and forty-one} \\
A: \entity{1941}}

\tiny{\setlength{\parindent}{0cm}
\textbf{id: \texttt{qw\_5050}} \\
Q: \textit{"Which Italian nobel prize winner (1934) wrote novels such as ""Mal Gioconda"" and switched to writing plays in 1910?"} \\
A: \entity{Pirandello} \\
A: \entity{Luigi Pirandello}}

\tiny{\setlength{\parindent}{0cm}
\textbf{id: \texttt{bt\_2403}} \\
Q: \textit{What was the name of the driver of the mail train robbed by the great train robbers} \\
A: \entity{Jack Mills (train driver)} \\
A: \entity{Jack Mills}}

\tiny{\setlength{\parindent}{0cm}
\textbf{id: \texttt{sfq\_2189}} \\
Q: \textit{What was the name of the private eye played by Trevor Eve on TV in the '70s?} \\
A: \entity{Shoestring (TV series)} \\
A: \entity{Eddie Shoestring}}

\tiny{\setlength{\parindent}{0cm}
\textbf{id: \texttt{qb\_5431}} \\
Q: \textit{Brazilian football legend Pele wore which number on his shirt?} \\
A: \entity{10} \\
A: \entity{ten}}

\tiny{\setlength{\parindent}{0cm}
\textbf{id: \texttt{qb\_4726}} \\
Q: \textit{Michael J Fox travels back to which year in the Wild West in the 1990 film ‘Back To The Future Part III’?} \\
A: \entity{one thousand, eight hundred and eighty-five} \\
A: \entity{1885}}

\tiny{\setlength{\parindent}{0cm}
\textbf{id: \texttt{odql\_8275}} \\
Q: \textit{Later a 'Blue Peter' presenter, who played 'Steven Taylor', an assistant to William Hartnell's 'Doctor Who'?} \\
A: \entity{PETER PURVES} \\
A: \entity{Peter Purves}}

\tiny{\setlength{\parindent}{0cm}
\textbf{id: \texttt{sfq\_962}} \\
Q: \textit{Which city was the subject of the 1949 song 'Dirty Old Town' by Ewan McColl?} \\
A: \entity{Salford} \\
A: \entity{Salford (disambiguation)}}

\tiny{\setlength{\parindent}{0cm}
\textbf{id: \texttt{tc\_1348}} \\
Q: \textit{In the late 60s Owen Finlay MacLaren pioneered what useful item for parents of small chldren?} \\
A: \entity{Baby Buggy} \\
A: \entity{Baby buggy}}

\tiny{\setlength{\parindent}{0cm}
\textbf{id: \texttt{qw\_12732}} \\
Q: \textit{General Franco, the Spanish military general, was head of state of Spain from October 1936 following the Spanish Civil War, until when?} \\
A: \entity{1975} \\
A: \entity{one thousand, nine hundred and seventy-five}}

\tiny{\setlength{\parindent}{0cm}
\textbf{id: \texttt{bt\_4495}} \\
Q: \textit{Which of the Great Train Robbers became a florist outside Waterloo station until he was found hanged in a lock up} \\
A: \entity{Buster Edwards} \\
A: \entity{Ronald \%22Buster\%22 Edwards}}

\tiny{\setlength{\parindent}{0cm}
\textbf{id: \texttt{sfq\_20394}} \\
Q: \textit{Which TV presenter, who died in February 2013, was for over 20 years the host of 'Mr \& Mrs'?} \\
A: \entity{Derek Batey} \\
A: \entity{Derek Beatty}}

\tiny{\setlength{\parindent}{0cm}
\textbf{id: \texttt{odql\_12918}} \\
Q: \textit{Which British political party leader is MP for Westmorland and Lonsdale?} \\
A: \entity{Tim Farron} \\
A: \entity{Timothy Farron}}

\tiny{\setlength{\parindent}{0cm}
\textbf{id: \texttt{odql\_13785}} \\
Q: \textit{Who wrote the lyrics for 'Sing', written to celebrate the Queen's Diamond Jubilee?} \\
A: \entity{Gary Barlow} \\
A: \entity{GARY BARLOW}}

\tiny{\setlength{\parindent}{0cm}
\textbf{id: \texttt{sfq\_20830}} \\
Q: \textit{Which top National Hunt trainer's establishment is based at Seven Barrows?} \\
A: \entity{NICKY HENDERSON} \\
A: \entity{Nicky Henderson}}

\tiny{\setlength{\parindent}{0cm}
\textbf{id: \texttt{wh\_2133}} \\
Q: \textit{Which T.V. Quiz show host used the catchphrase :- If its' up there, I'll give you the money myself ?} \\
A: \entity{LES DENNIS} \\
A: \entity{Les Dennis}}

\tiny{\setlength{\parindent}{0cm}
\textbf{id: \texttt{sfq\_15663}} \\
Q: \textit{The 27 episodes of which sitcom featuring Julia Mckenzie, Anton Rodgers and Ballard Berkley were first broadcast in the 1980s?} \\
A: \entity{Fresh Fields (TV series)} \\
A: \entity{Fresh Fields}}

\tiny{\setlength{\parindent}{0cm}
\textbf{id: \texttt{odql\_4871}} \\
Q: \textit{When US President James Garfield was shot in Washington DC in July 1881, what was he doing?} \\
A: \entity{WAITING FOR A TRAIN} \\
A: \entity{Waiting for a Train}}

\tiny{\setlength{\parindent}{0cm}
\textbf{id: \texttt{bb\_6592}} \\
Q: \textit{Which artist was born in Bradford in 1937?} \\
A: \entity{Hockney} \\
A: \entity{David Hockney}}

\tiny{\setlength{\parindent}{0cm}
\textbf{id: \texttt{qw\_10270}} \\
Q: \textit{Argentina invaded UK's Falkland Islands, Israel invaded Southern Lebanon, Canada became officially independent of the UK, Leonid Brezhnev, leader of the USSR, died, all in what year?} \\
A: \entity{one thousand, nine hundred and eighty-two} \\
A: \entity{1982}}

\tiny{\setlength{\parindent}{0cm}
\textbf{id: \texttt{bt\_4206}} \\
Q: \textit{Who was the first woman to be seen on Channel 4} \\
A: \entity{Carol Vorderman} \\
A: \entity{Carol Voderman}}

\tiny{\setlength{\parindent}{0cm}
\textbf{id: \texttt{qw\_8871}} \\
Q: \textit{Lieutenant General James Thomas Brudenell, who commanded the Light Brigade of the British Army during the Crimean War, was the 7th Earl of what?} \\
A: \entity{Cardigan} \\
A: \entity{Cardigan (disambiguation)}}

\tiny{\setlength{\parindent}{0cm}
\textbf{id: \texttt{sfq\_13639}} \\
Q: \textit{Which model village did Samuel Greg build to house workers at his nearby Quarry Bank Mill?} \\
A: \entity{Styal} \\
A: \entity{STYAL}}

\tiny{\setlength{\parindent}{0cm}
\textbf{id: \texttt{tc\_812}} \\
Q: \textit{Who was the defending champion when Martina Navratilova first won Wimbledon singles?} \\
A: \entity{Virginia Wade} \\
A: \entity{Sarah Virginia Wade}}

\tiny{\setlength{\parindent}{0cm}
\textbf{id: \texttt{sfq\_11790}} \\
Q: \textit{Opened in 1963, which London nightclub did Mark Birley name after his then wife?} \\
A: \entity{Annabel's} \\
A: \entity{ANNABELS}}

\tiny{\setlength{\parindent}{0cm}
\textbf{id: \texttt{qw\_8397}} \\
Q: \textit{In 1995, Steffi Graf became the only tennis player to have won each of the four grand slam events how many times?} \\
A: \entity{four} \\
A: \entity{4}}

\tiny{\setlength{\parindent}{0cm}
\textbf{id: \texttt{dpql\_3151}} \\
Q: \textit{On which river does Ipswich stand?} \\
A: \entity{Orwell (disambiguation)} \\
A: \entity{Orwell}}

\tiny{\setlength{\parindent}{0cm}
\textbf{id: \texttt{qw\_14634}} \\
Q: \textit{"Which Bob Dylan song begins ""You got a lotta nerveTo say you are my friend. When I was down, You just stood there grinning""?"} \\
A: \entity{Positively Fourth Street} \\
A: \entity{Positively 4th Street}}

\tiny{\setlength{\parindent}{0cm}
\textbf{id: \texttt{dpql\_1801}} \\
Q: \textit{Nick Begs was lead singer with which 80’s pop band?} \\
A: \entity{Kaja Googoo} \\
A: \entity{Kajagoogoo}}

\tiny{\setlength{\parindent}{0cm}
\textbf{id: \texttt{qw\_16011}} \\
Q: \textit{In 1483, who was appointed the first grand inquisitor of the Spanish Inquisition?} \\
A: \entity{Torquemada (disambiguation)} \\
A: \entity{Torquemada}}

\tiny{\setlength{\parindent}{0cm}
\textbf{id: \texttt{qw\_1933}} \\
Q: \textit{What remake of a British science-fiction serial broadcast by BBC Television in the summer of 1953 was staged live by BBC Four in 2005 with actors Jason Flemyng, Mark Gatiss, Andrew Tiernan, Indira Varma, David Tennant and Adrian Bower?} \\
A: \entity{Quatermass experiment} \\
A: \entity{The Quatermass Experiment}}

\tiny{\setlength{\parindent}{0cm}
\textbf{id: \texttt{odql\_2323}} \\
Q: \textit{Which 2009 film is a biopic of John Lennon?} \\
A: \entity{'NOWHERE BOY'} \\
A: \entity{Nowhere Boy}}

\tiny{\setlength{\parindent}{0cm}
\textbf{id: \texttt{bb\_522}} \\
Q: \textit{'The Battle of Trafalgar' is the work of which British painter?} \\
A: \entity{Joseph Turner} \\
A: \entity{Joseph Turner (disambiguation)}}

\tiny{\setlength{\parindent}{0cm}
\textbf{id: \texttt{qw\_4463}} \\
Q: \textit{Who discovered the two moons of Mars in 1877?} \\
A: \entity{Asaph Hall} \\
A: \entity{Asaph Hall III}}

\tiny{\setlength{\parindent}{0cm}
\textbf{id: \texttt{qz\_1111}} \\
Q: \textit{Which brand of beer does Homer Simpson drink regularly?} \\
A: \entity{Duff} \\
A: \entity{Duff (disambiguation)}}

\tiny{\setlength{\parindent}{0cm}
\textbf{id: \texttt{wh\_8}} \\
Q: \textit{In the novel 'Treasure Island' name the pirate shot dead by Jim Hawkins in the rigging of the Hispaniola} \\
A: \entity{Israel Hands} \\
A: \entity{ISRAEL HANDS}}

\tiny{\setlength{\parindent}{0cm}
\textbf{id: \texttt{qg\_3884}} \\
Q: \textit{Dow Constantine and Susan Hutchinson are currently running for was position?} \\
A: \entity{King County executive} \\
A: \entity{King County Executive}}

\tiny{\setlength{\parindent}{0cm}
\textbf{id: \texttt{sfq\_25940}} \\
Q: \textit{Which actor played the part of Ross Poldark in the BBC’s mid 1970’s television series?} \\
A: \entity{Robin Ellis} \\
A: \entity{ROBIN ELLIS}}

\tiny{\setlength{\parindent}{0cm}
\textbf{id: \texttt{odql\_12777}} \\
Q: \textit{For which 1960 film did Billy Wilder become the first person to win three Oscars for the same film?} \\
A: \entity{The Apartment} \\
A: \entity{The apartment}}

\tiny{\setlength{\parindent}{0cm}
\textbf{id: \texttt{bb\_4540}} \\
Q: \textit{Famous for 'Die Welt als Wille und Vorstellung', Arthur Schopenhauer (1788-1860) was a German?} \\
A: \entity{Philosophers} \\
A: \entity{Philosopher}}

\tiny{\setlength{\parindent}{0cm}
\textbf{id: \texttt{qb\_8589}} \\
Q: \textit{What is the nickname of the frontiersman Nathaniel Poe, played by Daniel Day Lewis, in the 1992, film ‘The Last of the Mohicans’?} \\
A: \entity{Hawkeye} \\
A: \entity{Hawkeye (disambiguation)}}

\tiny{\setlength{\parindent}{0cm}
\textbf{id: \texttt{bt\_2852}} \\
Q: \textit{Who played the part of Tina Seabrook in Casualty} \\
A: \entity{Claire Woodrow} \\
A: \entity{Claire Goose}}

\tiny{\setlength{\parindent}{0cm}
\textbf{id: \texttt{wh\_557}} \\
Q: \textit{Who duetted with Syd Owen on the single Better Believe It, which was released as part of the Children in Need appeal in 1995 ?} \\
A: \entity{PATSY PALMER} \\
A: \entity{Patsy Palmer}}

\tiny{\setlength{\parindent}{0cm}
\textbf{id: \texttt{odql\_3979}} \\
Q: \textit{What is the name of the character played by Nicole Kidman in the film 'Moulin Rouge'?} \\
A: \entity{Satine} \\
A: \entity{'SATINE'}}

\tiny{\setlength{\parindent}{0cm}
\textbf{id: \texttt{odql\_10365}} \\
Q: \textit{Which British girl won the Women's Junior Singles title at Wimbledon this year (2008)?} \\
A: \entity{LAURA ROBSON} \\
A: \entity{Laura Robson}}

\tiny{\setlength{\parindent}{0cm}
\textbf{id: \texttt{qf\_1735}} \\
Q: \textit{In what year did Elvis Presley and his parents move from Tupelo to Memphis?} \\
A: \entity{one thousand, nine hundred and forty-eight} \\
A: \entity{1948}}

\tiny{\setlength{\parindent}{0cm}
\textbf{id: \texttt{tc\_1468}} \\
Q: \textit{What was Pete Sampras seeded when he won his first US Open?} \\
A: \entity{twelve} \\
A: \entity{12}}

\tiny{\setlength{\parindent}{0cm}
\textbf{id: \texttt{qf\_2679}} \\
Q: \textit{Who on TV has played a scarecrow and a Time Lord?} \\
A: \entity{John Pertwee} \\
A: \entity{Jon Pertwee}}

\tiny{\setlength{\parindent}{0cm}
\textbf{id: \texttt{sfq\_11844}} \\
Q: \textit{In which year was Olaf Palme assassinated and the Chernobyl nuclear power station exploded?} \\
A: \entity{1986} \\
A: \entity{one thousand, nine hundred and eighty-six}}

\tiny{\setlength{\parindent}{0cm}
\textbf{id: \texttt{qf\_3578}} \\
Q: \textit{Cassandra was the pseudonym of which writer in the Daily Mirror?} \\
A: \entity{William Neil Connor} \\
A: \entity{William Connor}}

\tiny{\setlength{\parindent}{0cm}
\textbf{id: \texttt{qz\_3898}} \\
Q: \textit{How many times did Steffi Graf win the Ladies Singles at Wimbledon?} \\
A: \entity{seven} \\
A: \entity{7}}

\tiny{\setlength{\parindent}{0cm}
\textbf{id: \texttt{qw\_6782}} \\
Q: \textit{What is the disease that Stephen Hawking has?} \\
A: \entity{Motor neuron disease} \\
A: \entity{Motor neuron diseases}}

\tiny{\setlength{\parindent}{0cm}
\textbf{id: \texttt{qw\_15255}} \\
Q: \textit{"How many films were made by director Sir Peter Jackson from Tolkien's short book, ""The Hobbit""?"} \\
A: \entity{3} \\
A: \entity{three}}

\tiny{\setlength{\parindent}{0cm}
\textbf{id: \texttt{sfq\_7168}} \\
Q: \textit{Who invented the wind-up radio?} \\
A: \entity{Trevor Bayliss} \\
A: \entity{TREVOR BAYLISS}}

\tiny{\setlength{\parindent}{0cm}
\textbf{id: \texttt{sfq\_14433}} \\
Q: \textit{In Pride and Prejudice what was the first name of Mr Darcy?} \\
A: \entity{Fitzwilliam (disambiguation)} \\
A: \entity{Fitzwilliam}}

\tiny{\setlength{\parindent}{0cm}
\textbf{id: \texttt{odql\_8230}} \\
Q: \textit{Which single by 'Leapy Lee' reached number two in the UK charts in 1968?} \\
A: \entity{'LITTLE ARROWS'} \\
A: \entity{Little Arrows}}

\tiny{\setlength{\parindent}{0cm}
\textbf{id: \texttt{dpql\_1416}} \\
Q: \textit{Whose is the first tale in Chaucer’s Canterbury Tales?} \\
A: \entity{The Knight} \\
A: \entity{Knight (disambiguation)}}

\tiny{\setlength{\parindent}{0cm}
\textbf{id: \texttt{qw\_1672}} \\
Q: \textit{Which womens squash player won the World Open four times (1985, 1987, 1990 \& 1992) and the British Open eight times?} \\
A: \entity{Susan Devoy} \\
A: \entity{Susan Elizabeth Anne Devoy}}

\tiny{\setlength{\parindent}{0cm}
\textbf{id: \texttt{qz\_832}} \\
Q: \textit{Who wrote the novels About A Boy, How To Be Good and High Fidelity?} \\
A: \entity{Nick Hornby} \\
A: \entity{Hornby, Nick}}

\tiny{\setlength{\parindent}{0cm}
\textbf{id: \texttt{sfq\_22884}} \\
Q: \textit{Which TV series was about a pop group called 'Little Ladies' featuring Charlotte Cornwell, Julie Covington and Rula Lenska?} \\
A: \entity{Rock Follies} \\
A: \entity{Rock Follies of '77}}

\tiny{\setlength{\parindent}{0cm}
\textbf{id: \texttt{odql\_10746}} \\
Q: \textit{Who wrote the 1951 novel ‘The Caine Mutiny’?} \\
A: \entity{HERMAN WOUK} \\
A: \entity{Herman Wouk}}

\tiny{\setlength{\parindent}{0cm}
\textbf{id: \texttt{bb\_285}} \\
Q: \textit{Said to refer erroneously to the temperature at which book paper catches fire, the title of Ray Bradbury's 1953 novel about a futuristic society in which reading books is illegal, is called 'Fahrenheit...' what? 972; 451; 100; or 25?} \\
A: \entity{451} \\
A: \entity{four hundred and fifty-one}}

\tiny{\setlength{\parindent}{0cm}
\textbf{id: \texttt{odql\_7290}} \\
Q: \textit{Who was the driver of the limousine at the time of Diana Princess of Wales' death?} \\
A: \entity{HENRI PAUL} \\
A: \entity{Henri Paul}}

\tiny{\setlength{\parindent}{0cm}
\textbf{id: \texttt{sfq\_4368}} \\
Q: \textit{Which island in the Grenadines of St. Vincent was bought by Colin Tennant in 1958? Princess Margaret built a holiday home there in the 1960's.} \\
A: \entity{MUSTIQUE} \\
A: \entity{Mustique}}

\tiny{\setlength{\parindent}{0cm}
\textbf{id: \texttt{odql\_5476}} \\
Q: \textit{Which pop star had the real name of Ernest Evans?} \\
A: \entity{Chubby Checker} \\
A: \entity{'CHUBBY CHECKER'}}

\tiny{\setlength{\parindent}{0cm}
\textbf{id: \texttt{tc\_980}} \\
Q: \textit{"Which supermodel said, ""I look very scary in the mornings?"} \\
A: \entity{We don't wake up for less than \$10,000 a day} \\
A: \entity{Linda Evangelista}}

\clearpage



\end{document}